%% file: paper_template.tex
\documentclass[conference]{IEEEtran}
\usepackage{times}

\usepackage[numbers]{natbib}
\usepackage{multicol}
\usepackage[bookmarks=true,colorlinks]{hyperref}
\hypersetup{colorlinks,allcolors=[rgb]{0,0.07843,0.45098}}
\usepackage{xcolor}
\usepackage{amsmath}
\usepackage{bbm}
\usepackage{booktabs}
\usepackage{array}
\usepackage{xspace}
\usepackage{graphicx}
\usepackage{multirow}
\usepackage{pdfpages}
\usepackage{longtable} 
\usepackage{booktabs} 

\usepackage[export]{adjustbox}  
\usepackage{makecell}

\usepackage{listings}
\usepackage{xcolor}

\definecolor{codegreen}{rgb}{0,0.6,0}
\definecolor{codegray}{rgb}{0.5,0.5,0.5}
\definecolor{codepurple}{rgb}{0.58,0,0.82}
\definecolor{backcolour}{rgb}{0.95,0.95,0.92}

\lstdefinestyle{mystyle}{
    backgroundcolor=\color{backcolour},   
    commentstyle=\color{codegreen},
    keywordstyle=\color{magenta},
    numberstyle=\tiny\color{codegray},
    stringstyle=\color{codepurple},
    basicstyle=\ttfamily\footnotesize,
    breakatwhitespace=false,         
    breaklines=true,                 
    captionpos=b,                    
    keepspaces=true,                 
    numbers=left,                    
    numbersep=5pt,                  
    showspaces=false,                
    showstringspaces=false,
    showtabs=false,                  
    tabsize=2
}

\input{math_macros}

\newcommand{\myparagraph}[1]{\vspace{5pt}\noindent\textbf{#1}}

\newcolumntype{P}[1]{>{\centering\arraybackslash}p{#1}}
\newcolumntype{M}[1]{>{\centering\arraybackslash}m{#1}}

\newcommand{\model}{SetItUp\xspace}
\newcommand{\taskname}{ functional\xspace}

\newcommand{\objectSet}{\ensuremath{\mathcal{O}}}
\newcommand{\noise}{\ensuremath{\epsilon}}

\newcommand{\loss}{\ensuremath{\mathcal{L}}}

\makeatletter
\DeclareRobustCommand\onedot{\futurelet\@let@token\@onedot}
\def\@onedot{\ifx\@let@token.\else.\null\fi\xspace}

\def\eg{\emph{e.g}\onedot} 
\def\ie{\emph{i.e}\onedot}

\makeatother

\begin{document}

\title{``Set It Up!'': Functional Object Arrangement with Compositional Generative Models}




%
\author{\authorblockN{Yiqing Xu\authorrefmark{1},
Jiayuan Mao\authorrefmark{2},
Yilun Du\authorrefmark{2}, 
Tomas Loz\'{a}no-P\'{e}rez\authorrefmark{2},
Leslie Pack Kaelbling\authorrefmark{2},
and David Hsu\authorrefmark{1}}
\authorblockA{\authorrefmark{1}School of Computing,
Smart System Institute, National University of Singapore}
\authorblockA{\authorrefmark{2} CSAIL, Massachusetts Institute of Technology}
}

\maketitle

\begin{abstract}
    This paper studies the challenge of developing robots capable of understanding under-specified instructions for creating functional object arrangements, such as ``set up a dining table for two''; previous arrangement approaches have focused on much more explicit instructions, such as ``put object A on the table.''
    We introduce a framework, {\emph \model}, for learning to interpret under-specified instructions. \model takes a small number of training examples and a human-crafted program sketch to uncover arrangement rules for specific scene types.
    By leveraging an intermediate graph-like representation of {\emph{abstract spatial relationships}} among objects, \model decomposes the arrangement problem into two subproblems: i) learning the arrangement patterns from limited data and ii) grounding these abstract relationships into object poses.
    \model leverages large language models (LLMs) to propose the abstract spatial relationships among objects in novel scenes as the constraints to be satisfied; then, it composes a library of diffusion models associated with these abstract relationships to find object poses that satisfy the constraints.
    We validate our framework on a dataset comprising study desks, dining tables, and coffee tables, with the results showing superior performance in generating physically plausible, functional, and aesthetically pleasing object arrangements compared to existing models. \footnote{Project page: https://setitup-rss.github.io/}

\end{abstract}

\IEEEpeerreviewmaketitle
\input{sections/1-introduction}
\input{sections/3-method}
\input{sections/4-experiment}
\input{sections/2-related}
\input{sections/conclusion}

\section*{Acknowledgments}
This research is supported in part by the National Research Foundation (NRF), Singapore and DSO National Laboratories under the AI Singapore Program (AISG Award No: AISG2-RP-2020-016), NSF grant 2214177, AFOSR grant FA9550-22-1-0249, ONR MURI grant N00014-22-1-2740, and ARO grant W911NF-23-1-0034. Any opinions, findings and conclusions or recommendations expressed in this material are those of the author(s) and do not reflect the views of NRF Singapore.


\bibliographystyle{plainnat}
\bibliography{references}

\onecolumn
\appendix
\input{sections/appendix}

\end{document}

%% file: math_macros.tex
\usepackage{amsmath,amsfonts,bm}

\def\eqref#1{equation~\ref{#1}}

\def\1{\bm{1}}

\def\vg{{\bm{g}}}

\def\vp{{\bm{p}}}

\DeclareMathAlphabet{\mathsfit}{\encodingdefault}{\sfdefault}{m}{sl}
\SetMathAlphabet{\mathsfit}{bold}{\encodingdefault}{\sfdefault}{bx}{n}

\def\gD{{\mathcal{D}}}

\def\gG{{\mathcal{G}}}

\def\gO{{\mathcal{O}}}
\def\gP{{\mathcal{P}}}

\def\gR{{\mathcal{R}}}



%% file: sections/1-introduction.tex
\section{Introduction}
Developing robots capable of understanding human goals and making plans to achieve them is a crucial step forward in embodied intelligence. However, this endeavor is complicated by the inherent ambiguity and under-specification of a broad range of human goals. For example, consider one of the most common yet time-consuming daily tasks~\citep[American Time Use Survey;][]{tus}: setting and cleaning up tables. Human instructions are inherently under-specified; they can be as vague as ``Could you set up a dining table?'' Understanding such under-specified instructions is fundamentally challenging, requiring robots to understand and ground physical feasibility, object functionality, commonsense aesthetics, and user preferences.

\input{fig-texts/teaser}
\input{fig-texts/method-overview}

While a large body of work has tackled the problem of generating object arrangements at a table-top or even at a house scale, most of it focused on grounding spatial relationships and making robot plans under explicit and unambiguous instructions, \eg, ``put the plate next to the fork''~\citep{danielczuk2021object, driess2021learning, goodwin2022semantically, manuelli2019kpam, qureshi2021nerp, simeonov2021long, simeonov2023se, yuan2022sornet, zeng2021transporter}. By contrast, in this paper, we consider the task of generating object arrangements based on under-specified descriptions, such as ``{\it tidy up the study table},'' ``{\it set a Chinese dinner table for two},'' and ``{\it make space on a coffee table for a chess game}.'' Given such ambiguous instructions, our goal is to generate configurations of objects that are physically feasible, functional, aesthetic, and aligned with user preferences. We call this task {\it \taskname object arrangement} (FORM), and it presents three challenges. 

First, unlike many purely spatial relationships that have large-scale annotations~\cite {krishna2017visual} or can be synthetically generated according to human-written rules, global scene annotations for {\it \taskname} object arrangements are usually scarce. This data scarcity issue becomes more severe when considering the learning of subjective preferences of a particular user. Second, not only is there very little data available, but there is also a need to make arrangement plans for a wide variety of objects, many of which might be unseen during training. The criteria for successful {\it \taskname} object arrangement are not one-size-fits-all; they must apply to different objects and settings, leading to numerous acceptable arrangements, posing a significant challenge in generalization. Finally, pinpointing a universal measure or rules for the commonsense object arrangement is challenging due to its multifaceted nature. 

To tackle the under-specified and multifaceted nature of \taskname object arrangements, our first contribution is a novel task formulation and the corresponding evaluation metrics. For a given scene type, \eg, study desks or dining tables, we formalize the task as generating the poses of objects given their category names and shapes, based on a small set of examples.  We used 5 examples per scene type in our experiments. These training examples include paired instructions and reference object arrangements, which can be easily collected by users. Furthermore, to enable strong and controllable generalization to unseen instructions and objects, for each scene type, we provide a short program {\it sketch} written in Python-like domain-specific language as a hierarchical generative model specification. The sketch provides a list of function names, signatures, and descriptions, but does not include any implementation. Essentially, it decomposes the arrangement task into simpler subproblems as a global ``guideline'' for machine learning algorithms. Finally, our benchmark also comes with a collection of rule-based metrics and human experiment rubrics for holistically evaluating different solutions in terms of their physical feasibility, functionality, and other aesthetic aspects.

Our second contribution is a novel hierarchically generative approach, as illustrated in Figure~\ref{fig:teaser}a. Its key idea is to use a library of abstract spatial relationships, e.g., {\it left-of, horizontally-aligned}, as its intermediate prediction task. This breaks down the task of predicting object poses into two steps: generating a set of object relationships that should be satisfied in the final arrangement, and generating concrete object poses that comply with these relationships. In our system \model, we leverage large language models (LLMs) for the first prediction task and use a library of compositional generative models to ground the relationships into poses. 

This problem decomposition brought by our abstract relationship representation significantly improves the model along all evaluation metrics, especially its generalization to scenarios involving unseen objects and novel instructional contexts. On a new dataset we collected, which comprises three scene types: study desks, dining tables, and coffee tables, we compare our framework with methods based solely on neural generative models (inspired by~\citet{liu2022structdiffusion}) or large language models (inspired by~\citet{wu2023tidybot}). Both qualitative examples and quantitative human studies demonstrate that our model surpasses baselines in all aspects of generating physically plausible, functional, and aesthetically appealing object arrangement plans.

%% file: fig-texts/teaser.tex
\begin{figure}[tp]
    \centering
    \includegraphics[width=\columnwidth]{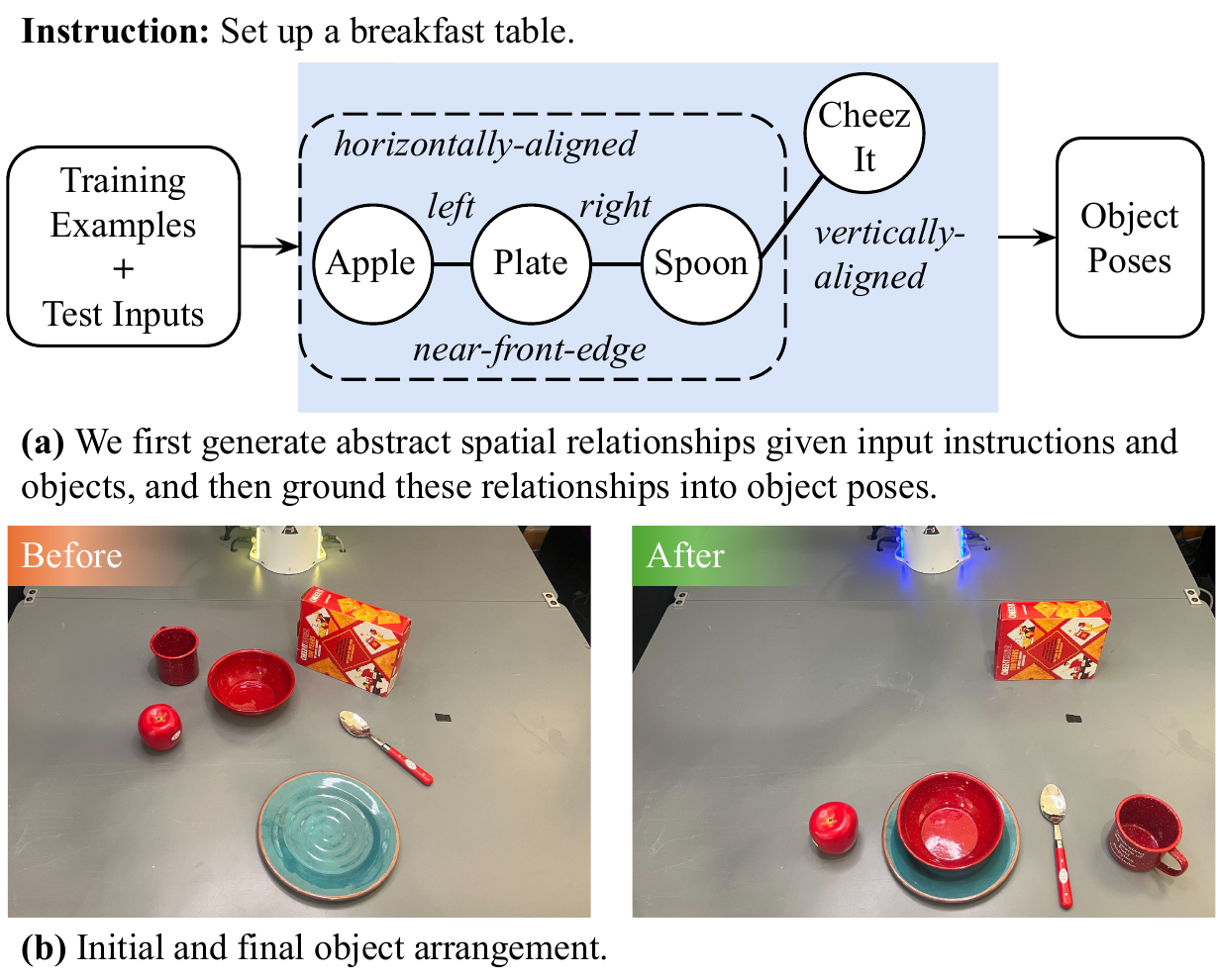}
    \vspace{-2em}
    \caption{(a) At test time, given human instruction and a set of objects (possibly unseen during training), our framework \model first generates a set of multi-ary spatial relationships among subsets of objects. These spatial relationships are based on a library of abstract spatial relationships and are visualized in the multi-ary graphical representation. (b) Then, we employ a compositional diffusion model to generate concrete object poses that a robot can execute based on a general motion planner.}
    \label{fig:teaser}
    \vspace{-1em}
\end{figure}

%% file: fig-texts/method-overview.tex
\begin{figure*}[tp]
    \centering
    \includegraphics[width=\linewidth, page=1]{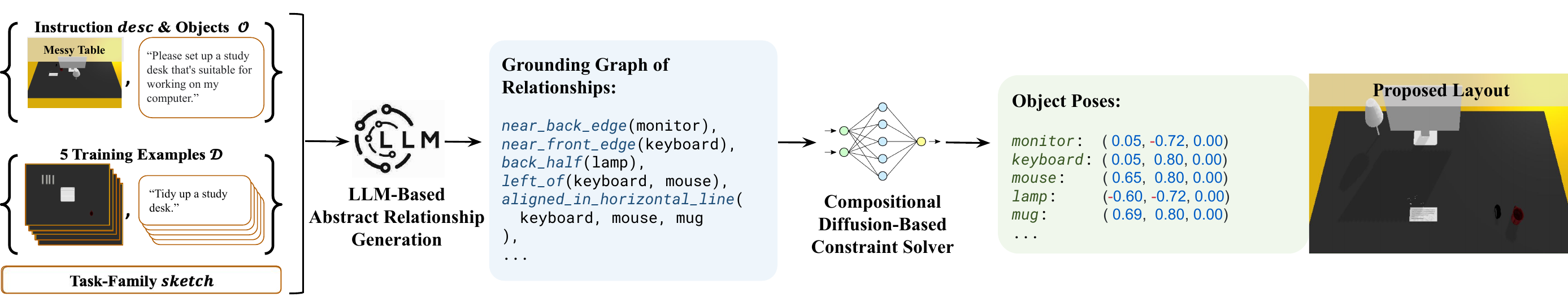}
    \caption{Overall architecture of \model. Given a novel instruction $\textit{desc}$ and a set of objects $\objectSet$, we first query an LLM to induce an abstract spatial relationship description of the target object arrangements. The input to the LLM also includes a handful of training examples $\gD$ and a human-defined task-family {\it sketch}. Next, we ground these abstract relationships into object poses by composing a library of diffusion models to generate object poses that simultaneously comply with all proposed spatial relationships.}
    \label{fig:model-overview}
\end{figure*}

%% file: sections/3-method.tex
\section{Problem Formulation}

We frame \taskname object arrangement (FORM), using a novel few-shot learning paradigm. Specifically, each object arrangement {\em task family} is a tuple $\langle \gD, \textit{desc}, \objectSet, \textit{sketch} \rangle$, where $\gD$ represents a small dataset of examples corresponding to a specific scene type and user preferences. Each training example $d \in \gD$ is a tuple $\langle \textit{desc}_d, \objectSet_d, \gP_d \rangle$. Here, $\textit{desc}_d$ denotes a natural language instruction for a specific task in that family (\eg, ``set up a dining table for two'') that implies a preferred object arrangement. $\objectSet_d$ and $\gP_d$ encode a desired scene configuration that fulfills the instruction $\textit{desc}_d$. Specifically, $\objectSet_d = \{o_0, o_1, \ldots, o_{N-1}\}$ denotes a collection of input objects categorized by two static properties: types ($n_i$) and shapes ($g_i$, represented as 2D bounding boxes in this paper). The desired ``output'' arrangement $\gP_d = \{ p_0, p_1, \ldots, p_{N-1} \}$ is represented by a planar pose for each object, expressed isn coordinates and orientation $(x, y, \theta)$\footnote{This basic problem formulation can be extended to 3D shapes and poses.}. Since we only focus on tabletop arrangement tasks in this paper, all poses are represented in a canonical table frame. We need very few (on the order of five) examples, making it easy for them to be collected by users.

At deployment time, we consider a new task instruction $\textit{desc}$ and a novel set of objects $\objectSet{} = \{o_0, o_1, \ldots, o_{N-1}\}$, which may vary in sizes and categories from the training examples; the system must determine a set of poses $\gP$ for each object. These poses should satisfy user instructions by being physically plausible, functional (logically positioned for their intended use), aesthetic (\eg, arrangements are visually appealing), and aligned with human preferences (conforming to the pattern of arrangements communicated in the training data $\gD$ as well as meet specific additional criteria in the user's instruction $\textit{desc}$).

To enable strong and controllable generalization to an unseen set of objects and instructions, we introduce a program {\it sketch} for each task family. This {\it sketch} defines a hierarchical generative model for generating object arrangements for a particular family of tasks. It uses a Python-like syntax to outline the task-solving procedures with only function names, signatures, and descriptions, but no implementations. Essentially, it decomposes the problem of generating the full arrangement into a hierarchy of smaller sub-problems. Each sub-problem has a simpler input-output specification, such as categorizing object types and generating the arrangement of a subset of the object. In this paper, we focus on leveraging human-defined sketch descriptions for three domains: {\it study desks, coffee tables, and dining tables}. Moving forward, the sketches could conceivably be generated from free-form language instructions from humans as in \citet{zelikman2023parsel}.

\section{Set It Up}

To address the challenges of data-efficient learning and robust generalization in \textit{\taskname} object arrangement (FORM) tasks, we introduce a novel hierarchical generative framework, namely \textit{\model}. The key idea of \model is to use a library of abstract spatial relationships to construct a grounding graph, which serves as an intermediate representation for solving the arrangement task. This library is composed of basic spatial relationships such as \textit{left-of} and \textit{horizontally-aligned}, which can be unambiguously defined by considering the geometric features of objects. Based on this library, we can decompose the problem of learning to generate object poses given their categories, shapes, and the task instructions into two subproblems: learning to generate the set of spatial relationships that encodes the arrangement plan, and learning to generate object poses based on a set of specified spatial relationships.

Figure~\ref{fig:model-overview} illustrates the overall framework. Central to our framework is the use of abstract spatial relationships as the intermediate representation. This high-level abstraction over object poses naturally breaks down FORM tasks into two generative sub-tasks. The first sub-task involves a LLM-based symbolic generative model that generates the functional abstract relationships for the test objects. The second sub-task involves a diffusion-based generative model that takes these abstract relationships as input and propose the corresponding object poses.
Essentially, \model combines large language models (LLMs) as a repository of commonsense knowledge and as a strong few-shot learner for generating abstract descriptions of the arrangement, and compositional diffusion models as powerful generative models to ground these spatial relationships into object poses. Given the task input tuple $\langle \gD, \textit{desc}, \gO, \textit{sketch} \rangle$, we first convert all training examples in $\gD$ into a language description based on the abstract spatial relationship library. Next, we prompt an LLM, based on the task inputs, to generate a set of abstract spatial relationships among test objects. This gives us a factor graph-like representation of the desired arrangement. Finally, we leverage a composition of the diffusion-based generative models associated with each abstract relationship to predict the output object poses associated with each object, which comply with the factor graph description generated by the LLM.

In the following sections, we will first present the library of abstract spatial relationships (Section~\ref{ssec:dsl}), where each relationship is associated with a simple geometric-rule-based classifier and a neural network-based generative model. Following that, we will introduce a compositional diffusion model inference algorithm capable of generating object poses based on the factor graph specification of object relationships. Finally, in Section~\ref{ssec:llm}, we specify our strategy for prompting an LLM based on the training examples, test inputs, and task-family sketches to propose the abstract spatial relationships.

\input{fig-texts/relationships}

\subsection{The Spatial Relationship Library}
\label{ssec:dsl}
Our spatial relationship library, $\gR$, encompasses 24 basic relationships listed in Table \ref{tab:relationships}. Each abstract relationship takes a (possibly variable-sized) set of objects as arguments --- and describes a desired spatial relationship among them. These abstract relationships provide one level of abstraction over 2D poses and, therefore, can serve as a natural input-output format for LLMs. On the other hand, they are sufficiently detailed to be directly interpreted as geometric concepts.  To extend the system to task families requiring novel relationships, it would be straightforward to augment this set.
Furthermore, these spatial constraints are defined unambiguously based on simple geometric transformations, which enables us to effectively classify and generate random object arrangements that satisfy a particular relationship. Finally and most importantly, this finite set of relationships can be {\it composed} to describe an expansive set of possible scene-level arrangements with indefinite number of objects and relationships.

Formally, each abstract spatial relationship $R$ is associated with two models: a classifier model $h_R$ and a generative model $f_R$. Let $k_R$ be the arity of the relationship. The classifier $h_R$ is a function, denoted as $h_R\left( g_1, \ldots, g_{k_R}, p_1, \ldots, p_{k_R} \right)$, that takes the static properties of $k_R$ objects (shapes represented by 2D bounding-boxes $\{ g_i \}$ in our case) and their poses $\{ p_i \}$ as input, and outputs a Boolean value indicating whether the input objects satisfy the relationship $R$. Similarly, the generative model $f_R$ is a function that takes the static properties of $k_R$ objects $\{ g_i \}$ and produces samples of $\{ p_i \}$ from the distribution\footnote{In practice, we always constrain the value range of $p_i$ to be within $[0, 1]^*$; therefore, this distribution is properly defined.} $q_R\left( p_1, \ldots, p_{k_R} \mid g_1, \ldots, g_{k_R} \right) \propto \mathbbm{1}\left[ h_R\left( g_1, \ldots, g_{k_R}, p_1, \ldots, p_{k_R} \right) \right]$, where $\mathbbm{1}[\cdot]$ is the indicator function. In general, $R$ (and therefore the associated $h_R$ and $f_R$) can be a set-based function and, therefore, may have a variable arity, in which case $k_R = \star$.

\myparagraph{Grounding graph.} Given the library of abstract spatial relationships $\gR$, we can encode a desired spatial arrangement of an object set $\gO$ as a graph of ground spatial relations, $\gG = \{r_i(o^i_1, \ldots, o^i_{k_i})\}_i$ where each $r_i \in \mathcal{R}$ is a relationship (such as {\it horizontally-aligned}), $k_i$ is the arity of $r_i$, and $o^i_1, \ldots, o^i_{k_i}$ are elements of $\mathcal{O}$. The objective is to produce a set of poses $\gP = \{ p_i \}$, such that, for all elements $(R, (o_1, \ldots, o_{k_R})) \in \mathcal{G}$, $h_R(g_1, \ldots g_{k_R}, p_1, \ldots, p_{k_R})$ is true, where $g_i$ is the corresponding shape of $o_i$.

We can interpret $(\gP, \gG)$ as a graph of constraints. Based on the probabilistic distribution specified with all relationships in $\gR$ (\ie, uniform distributions over allowable assignments), then $(\gP, \gG)$ can also be interpreted as a factor graph, specifying a joint distribution over values of $\gP$. This will be our inference-time objective.

\myparagraph{Classifying spatial relationships} Since all relationships used in our examples are unambiguously defined based on simple geometric transformations (\eg, by comparing the 2D coordinate of objects in a canonical table frame), we use a small set of rules to construct the classifier function $h_R$. We include the details of the rules in the appendix \ref{apped:abstract_rule}.  They could instead be learned in conjunction with the generative models for relationships that are defined through examples only. 

\input{fig-texts/single-diffusion-model}

\myparagraph{Generating spatial relationships.}
One straightforward approach to the overall problem would be to hand-specify $f_R$ for each relationship type and use standard non-linear optimization methods to search for $\gP$.
There are two substantial difficulties with this approach.
First, we may want to extend to relationship types for which we do not know an analytical form for $f$, and so would want to acquire it via learning.
Second, the optimization problem for sampling assignments to $P$ given a ground graph representation $(\gP, \gG)$ is highly non-convex and hence very difficult, and the standard method would typically require a great deal of tuning (e.g., of the steepness of the objective near the constraint boundary).

For these reasons, we adopt a strategy that is inspired by \citet{yang2023diffusion}, which is 1) to pre-train an individual diffusion-based generative model for each relationship {\em type} $R \in \mathcal{R}$ and 2) to combine the resulting ``denoising'' gradients to generate samples of $\gP$ from the high-scoring region.
One significant deviation from the method of \citet{yang2023diffusion} is that we train diffusion models for each relationship type completely independently and combine them only at inference time, using a novel inference mechanism that is both easy to implement and theoretically sound.

Our model architecture and the training paradigm are illustrated in Figure~\ref{fig:single-diffusion-model}. Specifically, for each relationship type $R \in \mathcal{R}$ of arity $k$, we require a training dataset of {\em positive examples} $\gD_R = \{(g_1, \ldots, g_k, p_1, \ldots, p_k)\}$ specifying satisfactory poses $\{p_i\}$ for the given object shapes $\{g_i\}$. Note that for set-based relations, the examples in the training set may have differing arity. For all relations in our library, we generate these datasets synthetically, as described in the appendix \ref{apped:abstract_rule}.

We construct a denoising diffusion model~\cite{ho2020denoising} for each relationship $R$ where the distribution $q_R(p_1, \ldots, p_k \mid g_1, \ldots, g_k)$ maximizes the likelihood $\{(g_1, \ldots, g_k, p_1, \ldots, p_k)\}$. We denote this distribution as $q_R(\vp \mid \vg)$ for brevity, where $\vp$ and $\vg$ are vector representations of the poses and the shapes, respectively. We learn a denoising function $\epsilon_R(\vp, \vg, t)$ which learns to denoise poses across a set of timesteps of $t$ in $\{1, \ldots, T\}$:
\begin{align*}  
\loss_{\textit{MSE}} = &\mathbbm{E}_{(\vp, \vg) \sim \mathcal{D}_R, \mathbf{\epsilon} \sim \mathcal{N}(\mathbf{0}, \mathbf{I}), t \sim U(0, T)} \\
&\left [ \left\|\mathbf{\noise} - \noise_{R}(\sqrt{\bar{\alpha}_t} \vp + \sqrt{1-\bar{\alpha}_t}\mathbf{\noise}, \vg, t) \right\|^2 \right ],
\end{align*}
where $t$ is a uniformly sampled diffusion step, $\mathbf{\epsilon}$ is a sample of  Gaussian noise, and $\bar{\alpha}_t$ is the diffusion denoising schedule.  In the case that $R$ has fixed arity, the network $\epsilon_\theta$ is a multi-layer perceptron (MLP); however, when it is set-based, we use a transformer to handle arbitrary input set sizes.  Details of these networks are provided in the appendix \ref{append:diffusion}.

The denoising functions $\{\epsilon_R(\vp, \vg, t)\}_{t=0:T}$ represent the score of a sequence of $T$ individual distributions, $\{q_R^t(\vp \mid \vg)\}_{t=0:T}$, transitioning from $q_R^0(\vp \mid \vg) = q_R(\vp \mid \vg)$ to $q_R^T(\vp \mid \vg) = \mathcal{N}(\mathbf{0}, \mathbf{I})$. 
Therefore, to draw samples with the diffusion process, we initialize a sample $\vp_T$ from $\mathcal{N}(\mathbf{0}, \mathbf{I})$ (\ie a sample from $q_R^T(\cdot)$). We then use a reverse diffusion transition kernel to construct a simple $\vp_{t-1}$ from distribution $q_R^{t-1}(\cdot)$, given a sample $p_t$ from $q_R^{t}(\cdot)$. This reverse diffusion kernel corresponds to:  
\begin{equation}
    \vp_{t-1} = B_t(\vp_t - C_t \epsilon_R(\vp_t, \vg, t) + D_t \mathbf{\xi} ), \quad \mathbf{\xi} \sim \mathcal{N}(\mathbf{0}, \mathbf{I})
    \label{eqn:reverse}
\end{equation}
where $B_t$, $C_t$, and $D_t$ are all constant terms and $\epsilon_R(\vp_t, \vg, t)$ is our learned denoising function. The final generated sample, $\vp_0$ corresponds to a sample from $q_R^0(\cdot) = q_R(\cdot)$ and is our final generated sample. 

\subsection{Pose Generation via Compositional Diffusion Models}
\label{ssec:diffusion}

In a single diffusion model, we can generate new samples from the learned distribution by sampling an initial pose $\vp_T \sim \mathcal{N}(\mathbf{0}, \mathbf{I})$, and then repeatedly applying the learned transition kernel, sequentially sampling objects $q_R^{t-1}(\cdot)$ until we reach $\vp_0$, which is a sample from the desired distribution.  

However, in the composed diffusion model setting, our target distribution is defined by an entire factor graph, and we wish to sample from a sequence of product distributions $\{\prod_{R \in G}  q_R^t(\vp \mid \vg)\}_{t=0:T}$ starting from an initial sample $\vp_T$ drawn from $\mathcal{N}(\mathbf{0}, \mathbf{I})$. For brevity, we refer to $\prod_{R \in G} q_R^t(\vp \mid \vg)$ as $q_{\textit{prod}}^t(\vp \mid \vg)$\footnote{Here we used a simplified notation. Each relation $g \in G$ is actually a tuple of $\left(R, (o_1, \ldots, o_{k_R})\right)$. It selects a particular relation $R$ and s subset of objects to which to apply it. Thus, the corresponding distribution $q^g$ should be defined as $q_R(\vp^g \mid \vg^g) = q_R(p_1, \ldots, p_{k_R} \mid g_1, \ldots, g_{k_R})$. For brevity, we used $\prod_{R\in G} q^R(\vp \mid \vg)$ to denote the composite distribution.}. While it is tempting to use the reverse diffusion kernel in Equation~\ref{eqn:reverse} to transition to each distribution  $q_{\textit{prod}}^t(\vp \mid \vg)$, we do not have access to the score function for this distribution~\cite{du2023reduce}.

We can instead transition between distributions by using annealed MCMC
, where we essentially use the composite score function $\sum_R \epsilon_R(\vp_t, \vg, t)$ across factors as a gradient for various MCMC transition kernels to transition from $q_{\textit{prod}}^t(\vp \mid \vg)$ to $q_{\textit{prod}}^{t-1}(\vp \mid \vg)$. \citet{du2023reduce} suggested a set of MCMC kernels to use in this process, which requires substantial extra work to find hyperparameters to enable accurate sampling.  We observe (and proved in Appendix~\ref{sect:reverse_ula}) that a very simple variant of the ULA sampler  
can be implemented by using the same parameters as the reverse kernel in Equation~\ref{eqn:reverse} on the composite score function
\begin{equation*}
    \vp_{t}' = \vp_t - C_t \sum_R \epsilon_R(\vp_t, \vg, t) + D_t \mathbf{\xi}, \quad \mathbf{\xi} \sim \mathcal{N}(\mathbf{0}, \mathbf{I}),
\end{equation*}
where $C_t$ and $D_t$ correspond to the same constant terms previously defined.
Note, however, that this is not the reverse sampling kernel needed to transition directly from $q_{\textit{prod}}^t(\vp \mid \vg)$  to $q_{\textit{prod}}^{t-1}(\vp \mid \vg)$ as this requires a different score function~\cite{du2023reduce}, but rather a MCMC sampling step for $q_{\textit{prod}}^t(\vp \mid \vg)$.

Our variant of the ULA sampler allows for a simple implementation for sampling from composed diffusion models. We can directly compute a composite score function and apply the reverse diffusion step with this score function at a fixed noise level as running one step of MCMC at a distribution $q_{\textit{prod}}^t(\cdot)$
Therefore, we can start with a random sample from $\mathcal{N}(\mathbf{0}, \mathbf{I})$ and then repeatedly run $M$ reverse diffusion step at each noise level to generate a final sample from $q_{\textit{prod}}^0(\cdot)$.

\subsection{Abstract Relationship Generation via Program Induction}
\label{ssec:llm}

\input{fig-texts/LLM-rule-induction}

Now that we have a module for going from a ground graph description into object poses, we further generate graph descriptions $\gG$ using a large language model in this section. As shown in Figure~\ref{fig:LLM-rule-induction}, our approach involves a two-stage generative process using the LLM. At the training stage, we synthesize a {\em task family rule-based program} that captures patterns for object arrangements. This program comprises detailed comments and code templates with unbound variables, as the specific objects and instructions are not yet defined. The synthesis process uses a small set of training examples, $\gD$, to instantiate a provided {\em sketch}. During the inference stage, given a new set of objects $\gO$ and a task instruction $\textit{desc}$, we bind the actual variables to the induced program, producing an {\em executable Python program}. This script is then executed to generate the ground relationship graph.

\input{fig-texts/sketch}

Recall that the training examples only record object poses, these numerical values cannot be effectively used by the LLM to infer abstract relationship patterns. Therefore, we need to ``translate" the pose information in training examples $\gD$ into a set of active abstract relationships. To do this, for each data point in the training set $(\textit{desc}_d, \gO_d, \gP_d) \in \gD$, we compute the set of primitive relationships that hold by applying the classifier $h_R$ for each abstract relationship type $R \in \gR$, to each subset of objects in $\gO_d$. We use a string encoding of all active relationships to describe each example scene.

\myparagraph{Program Sketch.} Our relationship generation procedure adopts a hierarchical program synthesis framework. It is based on a human-defined, task-family-specific {\it sketch}, with an example shown in Figure \ref{fig:sketch}. Each sketch includes several functions with meaningful names, signatures, and descriptions, organized in a sequence that outlines the steps to solve the complex task. The sketch is crucial for generating abstract relationships. It acts as a guide that helps the LLM generate object relationships step-by-step. With this strong guidance, the LLM can create task-family-specific rule-based program from just five examples. Instead of directly regressing the final relationship proposals on these five training examples, we use the examples mainly to instantiate the subroutines in the sketch and to write detailed comments. Such high-level sketches decompose the generation task into four types of subroutines: 1) extracting the task relevant information from the instruction (\eg, number of diners), 2) categorizing objects into groups (\eg, finding all input devices to a computer), 3) generating arrangements for the objects within each group(\eg, all input devices), and 4) generating arrangements for objects among groups. These subroutines usually require only a single ``step'' of reasoning, such as directly extracting a number from the instruction or generating object arrangements for a smaller subset of the objects. In this work, we represent the sketch in Python-like syntax, but in future work, it would be important, like in \citet{zelikman2023parsel}, to accept sketches in natural language and convert them into Python or a simpler language such as their Parsel. In practice, we follow the four subroutines and manually decompose the prediction task into 7-10 functions; we provide the sketches for the three task families we study in this work in the appendix \ref{ssec:prompt}. To add a new task family, such as setting up for a Scrabble game, it is only necessary to provide a sketch and a few example set-ups. Furthermore, it is crucial to note that this high-level sketch does not contain concrete implementations for any subroutines; we will prompt the LLM, given the training examples, to solve each subproblem.

\myparagraph{Program Induction.}
Given the sketch and training examples, we query the LLM to generate a rule-based program. This program comprising rules and patterns for each subproblem, but with unbound variables (i.e., the actual objects and instructions are not yet specified). Specifically, the program encapsulates the rules associated with each function in the form of code comments and templates. For example, it may include instructions for identifying the key object in a study desk setup task. During the rule induction stage, we input the textual descriptions of the training examples, which are detailed in terms of their spatial relations, to the LLM. We then prompt the LLM to summarize the patterns in these examples into either comments or code templates with unbound variables. After the rule induction stage, we obtain a program with comments that summarizes the patterns derived from the training examples. Examples of a sketch and a resulting LLM-generated program are illustrated in Figure~\ref{fig:sketch}a and b.

\myparagraph{Variable Binding and Program Execution.} 
At performance time, given a new set of objects $\gO_d$, task description $\textit{descr}$, we first use the LLM to bind these variables to the induced program. This process incorporates the contextual information to generate an executable Python program. Then, we execute the program to return the set of functional abstract spatial relationships. Specifically, we prompt the LLM to read the test instructions $\textit{desc}$ and the object names in $\gO$, and fill in scene-specific details for each program. An example of an executable program with variable bindings, \verb|extract_main_devices|, is shown in Figure~\ref{fig:sketch}c:  the LLM generates a list of key objects for the new scene and incorporates it into a return statement. To generate the final set of relationships among objects, we simply execute this program with variable bindings, which now include the scene-specific comments and implementation, as well as the task description to the LLM, to generate the list of relationships.

Empirically, we have found that LLM-generated relationships often suffer from inconsistency or incompleteness. Inconsistency arises when the LLM proposes several relationships that cannot be simultaneously satisfied. Incompleteness emerges when the LLM does not provide sufficient information to predict the placement of a target object. To address these issues, we employ an iterative self-reflection process to refine the initial programs generated by the LLM. At each refinement iteration, the input to the LLM includes the task inputs, the fully instantiated programs generated by the rule instantiation step, and its execution result, which is a set of abstract spatial relationships. We then instruct the LLMs to compose two language summaries: one describing the desired scene configuration based only on the task inputs, and the other describing the scene configuration as inferred from the program execution results. Next, we prompt LLM to identify inconsistency and incompleteness between two language summaries and fix the instantiated setup rules. We include all detailed instructions and prompts to the LLM in the appendix \ref{ssec:prompt}.

%% file: fig-texts/relationships.tex
\begin{table}[tp]
\centering
\caption{The set of abstract spatial relationships among objects. These relationships are formally defined by rules based on 2D object shapes and poses in the camera frame.}
\label{tab:relationships}
\setlength{\tabcolsep}{4.3pt}
\vspace{-0.5em}
\begin{tabular}{ll}
\toprule
\multicolumn{2}{l}{\textbf{Unary Relationships}}  \\ \midrule
central\_column & central\_row \\
central\_table & left\_half \\
right\_half & front\_half \\
back\_half & near\_left\_edge \\
near\_right\_edge & near\_front\_edge \\
near\_back\_edge &  \\ \midrule
\multicolumn{2}{l}{\textbf{Binary Relationships}} \\ \midrule
horizontally\_aligned & vertically\_aligned \\ 
horizontal\_symmetry\_on\_table & vertical\_symmetry\_on\_table \\ 
left\_of & right\_of \\ 
centered & on\_top\_of \\ \midrule
\multicolumn{2}{l}{\textbf{Ternary Relationships}} \\ \midrule
horizontal\_symmetry\_about\_axis\_obj & vertical\_symmetry\_about\_axis\_obj \\ \midrule
\multicolumn{2}{l}{\textbf{Variable-Arity Relationships}} \\ \midrule
aligned\_in\_horizontal\_line & aligned\_in\_vertical\_line \\ 
regular\_grid &  \\ \bottomrule
\end{tabular}
\vspace{-1.5em}
\end{table}

%% file: fig-texts/single-diffusion-model.tex
\begin{figure}[tp]
    \centering
    \includegraphics[width=\linewidth]{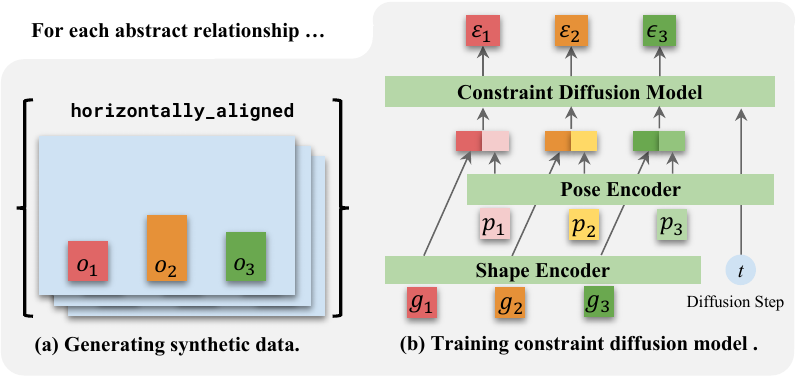}
    \caption{Training a single constraint diffusion model involves a two-stage process. First, for every abstract relationship listed in Table \ref{tab:relationships}, we generate a synthetic dataset based on predefined rules. Then, we train a relation-specific diffusion model that can draw samples of object poses that satisfy the relationship.}
    \label{fig:single-diffusion-model}
\end{figure}

%% file: fig-texts/LLM-rule-induction.tex
\begin{figure}[tp]
    \centering
    \includegraphics[width=\linewidth]{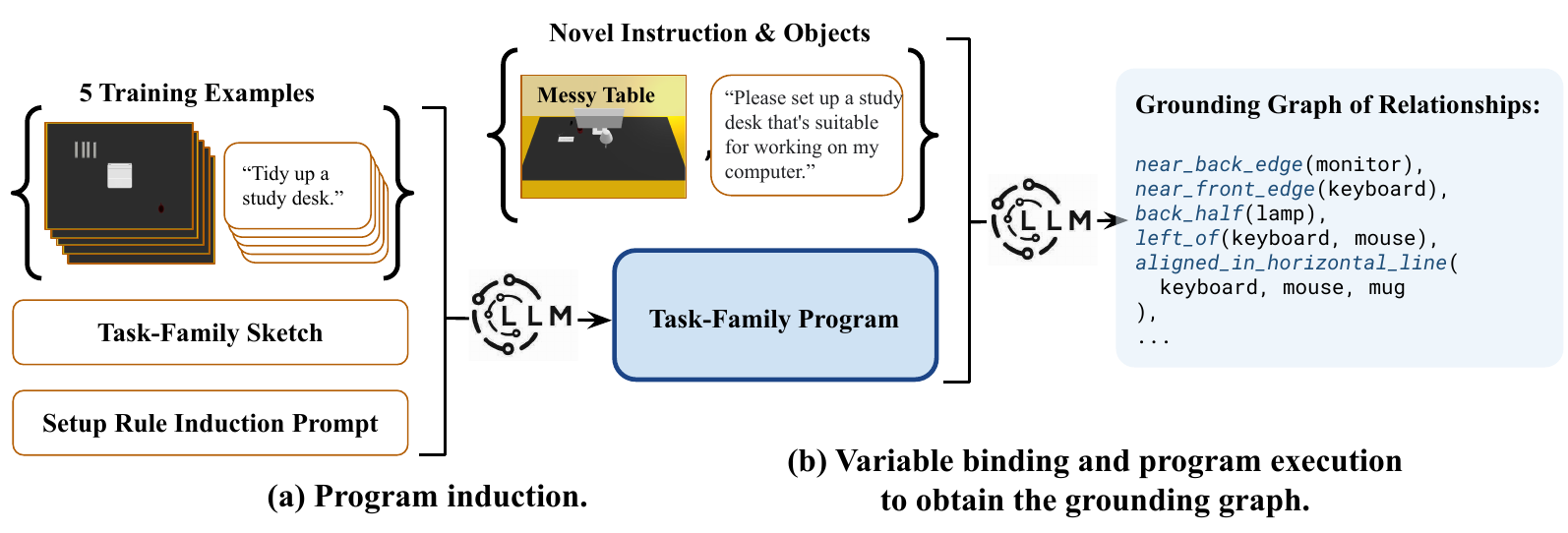}
    \caption{Abstract relationship generation through rule induction involves two phases. Initially, in the program induction phase, we employ an LLM to create a ``setup'' rule-based program from a few training examples and a high-level task-family sketch defined by humans. This program contains rules and patterns for various subproblems, but it has unbound variables (i.e., the actual objects and instructions are not specified yet). In the second phase, with a new instruction and a list of test objects, the LLM binds these variables to the induced program to create an executable Python program. This executable Python program is then used to generate the final set of abstract spatial relationships as a ground graph.}
    \label{fig:LLM-rule-induction}
\end{figure}

%% file: fig-texts/sketch.tex
\begin{figure}[tp]
 \centering
    \includegraphics[width=\columnwidth, page=1]{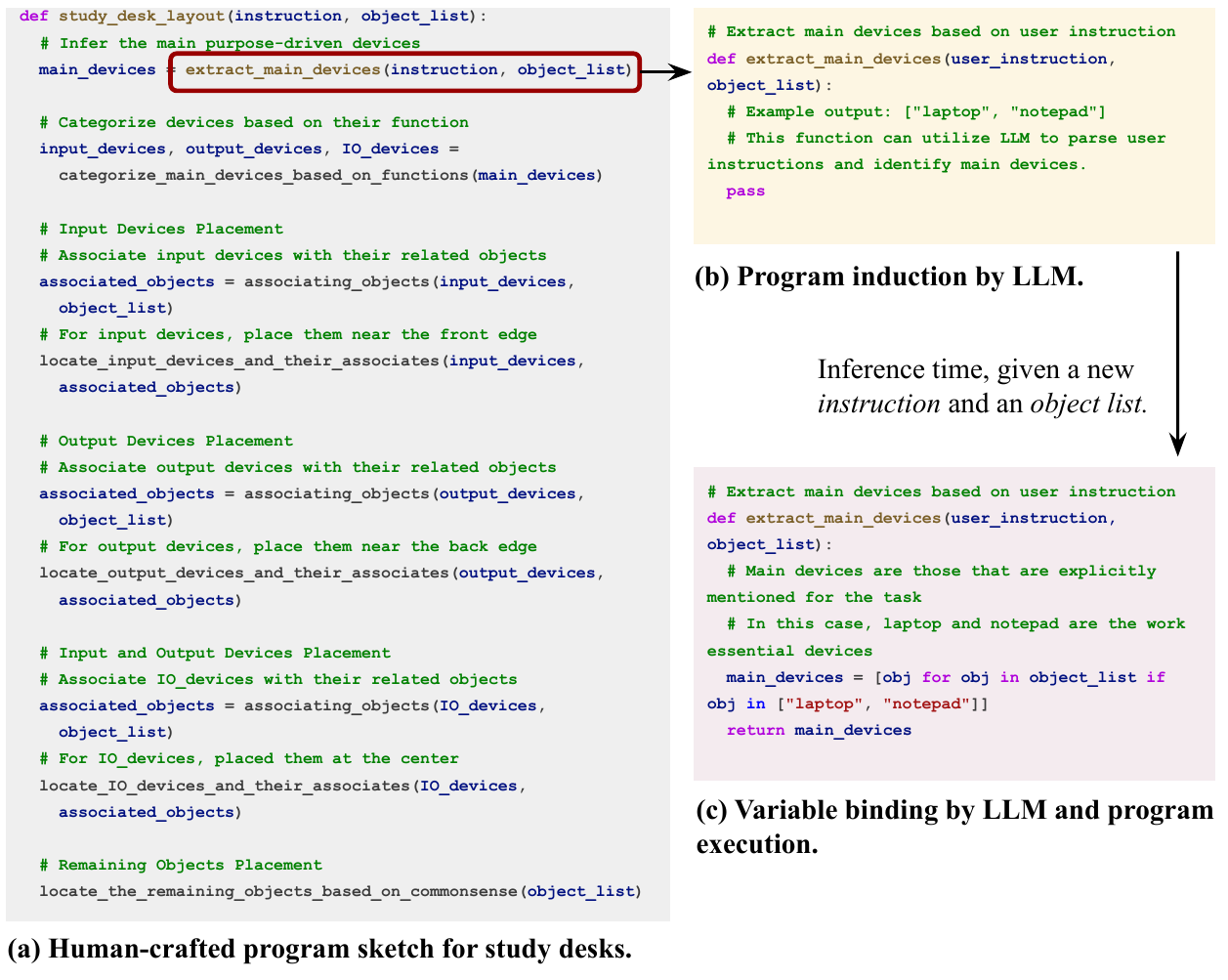}
\caption{Example process of using an LLM to instantiate a program sketch. Sub-figure (a) presents an example of the initial program sketch. We provide this program sketch, along with five training instances, to the LLM. The LLM then creates a rule-based program, summarizing the common patterns in the form of code comments and/or templates, but with unbound variables, as illustrated in (b). Finally, given new objects and instructions, the LLM binds these variables to the induced program and generates an executable Python program. This program is then used to generate the object grounding graph. An example of an executable Python program with variable bindings during inference time is depicted in (c).}
\label{fig:sketch}
\end{figure}

%% file: sections/4-experiment.tex
\section{Experiments}

\input{fig-texts/experiment-intro}

We evaluated our method in three scene types: study desks, dining tables, and coffee tables. They involve different types of objects, different aesthetic patterns, and different types of human needs (\eg, using a laptop vs. paper and pencil, formal dining vs. casual dining). They also progressively increase in scene complexity: the study desk has the fewest objects and important relationships, and the dining table features the most. We generated each scene with household items that commonly appear in the particular scene type. There are in total 15 distinct tasks created for each environment: 5 for training and 10 for evaluation. We include the training and testing examples in the appendix \ref{append:task_family}. Figure~\ref{fig:experiment-intro} shows examples of each scene type and the final object arrangements generated by \model.

We compare \model to several baselines according to the following criteria: How do different models perform across various aspects of the arrangement task? How does our neuro-symbolic design compare with monolithic neural networks or straightforward large language model predictions? Finally, how do different models generalize to novel scenarios involving unseen instructions and objects?

\subsection{Baselines}
We have implemented two baselines and an ablation variation of our model. Implementation details are included in the appendix \ref{append:baseline_implementation}.

{\noindent \bf End-to-End diffusion model}: we implement an end-to-end diffusion model for language-conditioned object arrangement, inspired by StructDiffusion \cite{liu2022structdiffusion}. We trained the model using a combination of the same synthetic data and 15 tidy scene examples (5 per scene type) as our model. At test time, it directly generates object poses based on the language instructions and object shapes.

{\noindent \bf Direct LLM Prediction}: This baseline directly leverages a large language model (LLM) to predict test object poses, inspired by Tidybot \cite{wu2023tidybot}. It is conditioned on 15 given examples (5 for each scene type), as well as the object categories and shapes.

{\noindent \bf LLM-Diffusion}: This model is a simplified variant of our approach, omitting the program sketch, rule induction, and iterative self-reflection mechanisms. It directly prompts an LLM to generate abstract object relationships and employs the same diffusion-model-based inference for grounding.

\subsection{Evaluation Metrics}

A {\it \taskname object arrangement} must meet several criteria: physical feasibility (i.e., being collision-free), functionality (i.e., serving the intended purpose as per human instructions), and overall convenience and aesthetic appeal (i.e., neat arrangement with alignments and symmetries). While physical feasibility and functionality can be objectively evaluated using rule-based scripts, the subjective nature of convenience and aesthetic appeal precludes such fixed evaluations. Therefore, we employ both rule-based auto-grading and human experiments to evaluate models.

\myparagraph{Physical feasibility:} We define the physical feasibility score of an arrangement as the proportion of collision-free objects in the final layout using a 2D collision detector. 

\myparagraph{Basic functionality:} This measures the proportion of functional relationships satisfied in the final scene configuration. For each scene, we use a set of manually defined rules to generate the basic functional relationships, including constraints that key objects (\eg, the utensil in a dining table setting) are within reach of the user (The user position(s) is predetermined), unobstructed, and arranged to fulfill their intended function (\eg, left-handedness).

\myparagraph{Aesthetics, convenience, and other preferences:} We conduct a human evaluation with 20 graduate students. Each participant evaluates 30 scenes generated by either our method or a baseline, assigning a score from 1 to 5 based on a set of criteria. Table \ref{tab:grading_guidelines} presents the grading criteria. We report the average score for each method across every scene type. For ease of evaluation, we rendered the 2D object poses using PyBullet~\citep{pybullet}, a physical simulator. It is important to note that the 3D models of the objects were not used in any of the methods for proposing object poses.

\input{fig-texts/grading_guidelines}

\input{fig-texts/combined_comparison}

\subsection{Results}
Table~\ref{tab:combined_comparison} shows the overall performance of all models across three scene types and three evaluation metrics.  Overall, our model \model consistently outperforms all baselines, achieving a nearly perfect collision-free rate (indicating physical feasibility) and recovering over 80\% of the relationships related to basic functionalities. Additionally, it consistently scores high across all human evaluations, indicating its success in achieving functional and aesthetically pleasing arrangements. The direct LLM prediction baseline and our model variant (LLM-Diffusion) exhibit similar rankings across all three evaluation metrics. We explore the behavior of the direct LLM prediction baseline and compare it with \model in detail in the remaining part of this section. The LLM-Diffusion baseline shows slightly better performance than the direct LLM prediction baseline in tasks with fewer objects and constraints (i.e., the study desk) and underperforms in tasks with many objects (i.e., the dining table). This implies that our approach of program-structured prompting and self-reflection-based refinement improves the overall system performance. A notable failure mode of the LLM-Diffusion variant is its frequent generation of conflicting relationships among objects, subsequently leading to collisions in the final scene configuration, and therefore, resulting in low human evaluation scores. The End-to-End diffusion model is the worst-performing method across all metrics and scene types, primarily due to its inability to leverage additional commonsense knowledge (e.g., from LLMs) during inference. This results in poor generalization to new tasks with only a limited number of training examples. We will further examine its generalization behavior in Section~\ref{sec:generalization}.

\input{fig-texts/visual-comparison}

Figure~\ref{fig:visual-comparison} illustrates the qualitative scene arrangements generated by different methods across all three scene types. By comparing the behavior of various methods, we aim to derive insights into the effectiveness of abstract relationships and compositional generative models.

\myparagraph{Direct LLM prediction vs. Ours: the importance of abstract relationships.} A notable observation in Table~\ref{tab:combined_comparison} is that the direct LLM prediction baseline typically achieves a higher physical feasibility score compared to its functionality score. This pattern becomes more apparent when compared to the LLM-Diffusion variant of our model. By contrast, the LLM-Diffusion variant maintains a consistent functionality score, even though its physical feasibility score decreases with an increase in the number of objects. Examining the scene configurations in Figure~\ref{fig:visual-comparison} more closely, we observe that the direct LLM prediction baseline excels at creating scenes where objects are well-separated and aesthetically arranged but often fails to fulfill functional requirements. For instance, in the dining table task, while all plates and utensils are neatly placed, the setup does not functionally accommodate two diners. This underscores the significance of leveraging abstract relationships as intermediate interfaces between large language models and the physical scene; it helps the model to better reason about important spatial relationships for the specific task.

\myparagraph{End-to-End diffusion model vs Ours: the importance of compositional diffusion models.}
Let's now compare the performance of the end-to-end diffusion model with our framework. It's important to note that both models are trained on the same dataset, including 30,000 examples of synthetically generated single-relationship arrangements and 15 scenes of human-labeled tidy arrangements. Looking at the dining table task in Figure~\ref{fig:visual-comparison}, we find that the end-to-end diffusion model manages to generalize to this novel test scene to a certain extent but fails to place all objects in a physically feasible manner. This suggests that it struggles with reusing its training data on individual relationship types while generating the global scene arrangements.

A critical insight into the problem of \textit{tidy object arrangement} is that generating synthetic data for a single relationship type is relatively inexpensive, whereas there is a significant lack of scene-level annotations of paired instructions and object arrangements. As a result, a compositional learning framework is preferred over a monolithic one, as it can learn individual relationships from synthetically generated single-relationship examples and then compose them at test time. This explicit compositional structure demonstrates a stronger performance compared to the end-to-end diffusion baseline, which naively mixes training data for single relationships and scene-level arrangements directly.

\subsection{Generalization}
\label{sec:generalization}

\input{fig-texts/in_vs_out_dist}

We further break down our quantitative results to analyze generalization across different dimensions of the problem: generalization to novel instructions, as well as to larger scenes with more objects and relationships.

\myparagraph{Generalization to novel instructions.}
Recall that our dataset comprises 5 training examples and 10 test examples for each scene type. We manually label each test example as either a ``seen instruction'' (i.e., it is similar to an instruction in the training examples) or a ``novel instruction.'' We then compute the average human evaluation score for both groups of instructions. Figure~\ref{fig:in_vs_out_dist} presents the results.

Our \model achieved the highest scores in both categories, with no significant performance drop when generalizing to novel instructions. By contrast, the end-to-end diffusion model demonstrated the weakest generalization. Its monolithic model structure, heavily reliant on the coverage of training data, limits its generalization to new instructions. The direct LLM prediction baseline and our ablation variant experienced similar performance drops on novel instructions for study desks and coffee tables. We attribute the direct LLM prediction baseline's poor generalization to novel instructions to its limited capability to reason about important functional relationships within the scene, as discussed in the previous section. Our full model also surpasses the LLM-Diffusion baseline, highlighting the effectiveness of our rule induction steps and self-reflection-based refinements for improved consistency and completeness.
\input{fig-texts/scalability}

\myparagraph{Generalization to more objects and complex scenes.}
Figure~\ref{fig:scalability} illustrates how the human evaluation score varies with different numbers of objects and functional relationships. Our \model (represented by dark red dots) achieves high scores consistently across the spectrum of scene complexity. By contrast, we see a noticeable performance drop for all other methods as the scene becomes more complex.

We believe that the strong scalability of our model comes from two important designs of the system. First, in the abstract relationship generation stage, our rule induction step can infer abstract rules about arrangement patterns and, therefore, generalize better to larger scenes. Second, the compositional design of our pose diffusion models allows for aggressive generalization to scenes with a greater number of objects and relationships, due to the composition achieved by explicitly summing up individually trained energy functions. This is consistent with the findings from \citet{yang2023diffusion} on how the compositional diffusion model compared to monolithic models in terms of generalization.

%% file: fig-texts/experiment-intro.tex
\begin{figure}[tp]
    \centering
    \includegraphics[width=\linewidth]{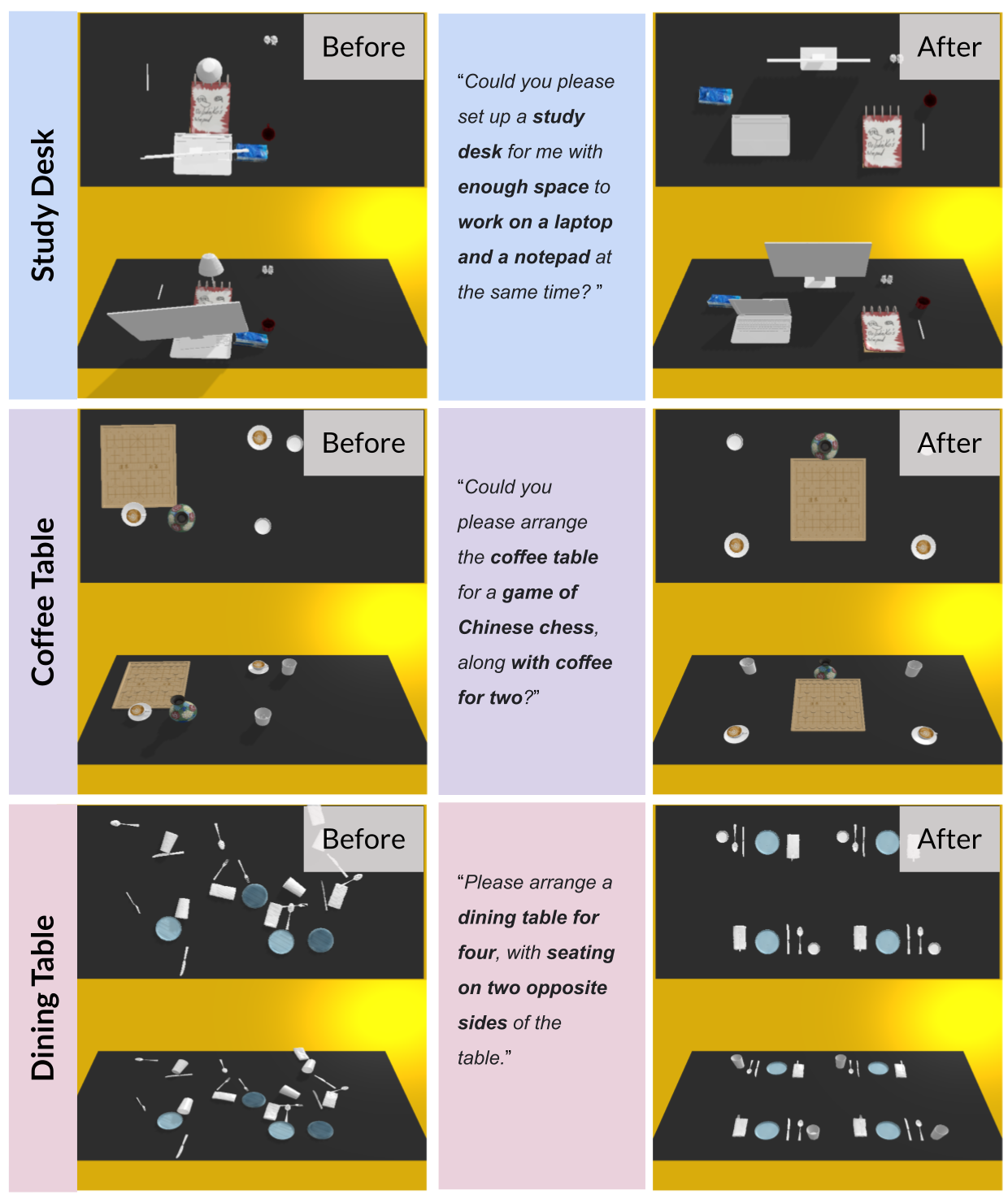}
    \caption{Illustration of three scene types (before and after). The after-scene configuration is generated by \model.}
    \label{fig:experiment-intro}
\end{figure}

%% file: fig-texts/grading_guidelines.tex
\begin{table}[tp]
\centering
\caption{Grading guidelines for human evaluation.}
\begin{tabular}{c m{7cm}}
\toprule
{\bf Points} & {\bf Grading Criteria} \\ \midrule 
{\bf 1} & Functionally inadequate. The setup does not serve the intended purpose in the user instruction.\\ \midrule 
{\bf 2} & Partial functional but inconvenient. The setup can somewhat serve the intended purpose, but the arrangement makes the intended activity inconvenient since it does not conform to the user's habits or social norms. \\ \midrule 

{\bf 3} & Functional, but cluttered. Key items are arranged properly to serve the intended purpose, but the space is overcrowded and lacks efficient organization.\\ \midrule 

{\bf 4} & Fully functional, lacks aesthetic appeal. Functionality and accessibility are good, but the arrangement lacks in visual harmony and alignment. \\ \midrule 

{\bf 5} & Fully functional, and aesthetically pleasing. All items are well-placed, easy to access, and efficiently organized. Furthermore, the setup is visually appealing with proper alignment and symmetry.gra\\ \bottomrule 
\end{tabular}
\label{tab:grading_guidelines}
\end{table}

%% file: fig-texts/combined_comparison.tex
\begin{table*}[tp]
\centering
\caption{Holistic evaluation of different methods across task families.}
\label{tab:combined_comparison}
\setlength{\tabcolsep}{4.5pt}
\begin{tabular}{@{}lcccccccccc@{}}
\toprule
\multirow{2}{*}{\textbf{Model}} & \multicolumn{3}{c}{\textbf{Physical Feasibility (\%)}} & \multicolumn{3}{c}{\textbf{Functionality (\%)}} & \multicolumn{3}{c}{\textbf{Overall Human Judgement (1-5)}} \\ \cmidrule(l){2-10} 
                                & Study & Coffee & Dining & Study & Coffee & Dining & Study & Coffee & Dining \\ \midrule
End-to-End Diffusion Model   &  37.7$_{\pm 32.1}$          & 39.6$_{\pm 17.3}$  &  7.03$_{\pm 9.59}$   &    49.8$_{\pm 21.5}$       &   53.5$_{\pm 13.1}$     & 37.5$_{\pm 11.1}$    & 1.89$_{\pm 0.834}$      &   1.91$_{\pm 0.650}$  &   1.58$_{\pm 0.433}$ \\
Direct LLM     &  63.1$_{\pm 27.3}$  &  58.6$_{\pm 22.5}$ &  63.6$_{\pm 20.1}$  &   59.7$_{\pm 11.7}$    &  53.7$_{\pm 22.3}$   & 56.4$_{\pm 15.3}$  &  3.01$_{\pm 0.901}$        &   2.84$_{\pm 1.060}$        &    2.33$_{\pm 0.416}$  \\ \midrule
LLM-Diffusion (Our ablation)   &  69.4$_{\pm 19.2}$   &  41.0$_{\pm 20.1}$       &   26.7$_{\pm 13.8}$   &  69.8$_{\pm 11.6}$   &   44.3$_{\pm 10.4}$      &    46.5$_{\pm 17.8}$  & 3.67$_{\pm 0.934}$      &  3.16$_{\pm 0.657}$       &   1.87$_{\pm 0.652}$      \\
\model (Ours)   &  {\bf 95.0}$_{\pm 10.0}$   & {\bf 98.1}$_{\pm 3.83}$             &  {\bf 95.8}$_{\pm 6.72}$  & {\bf 94.1}$_{\pm 6.04}$  &  {\bf 84.4}$_{\pm 13.5}$      &    {\bf 91.5}$_{\pm 13.8}$    & {\bf 4.49}$_{\pm 0.343}$     & {\bf 4.47}$_{\pm 0.211}$        &   {\bf 4.79}$_{\pm 0.190}$     \\ \bottomrule
\end{tabular}
\end{table*}

%% file: fig-texts/visual-comparison.tex
\begin{figure*}[tp]
    \centering
    \includegraphics[width=\linewidth, page=1]{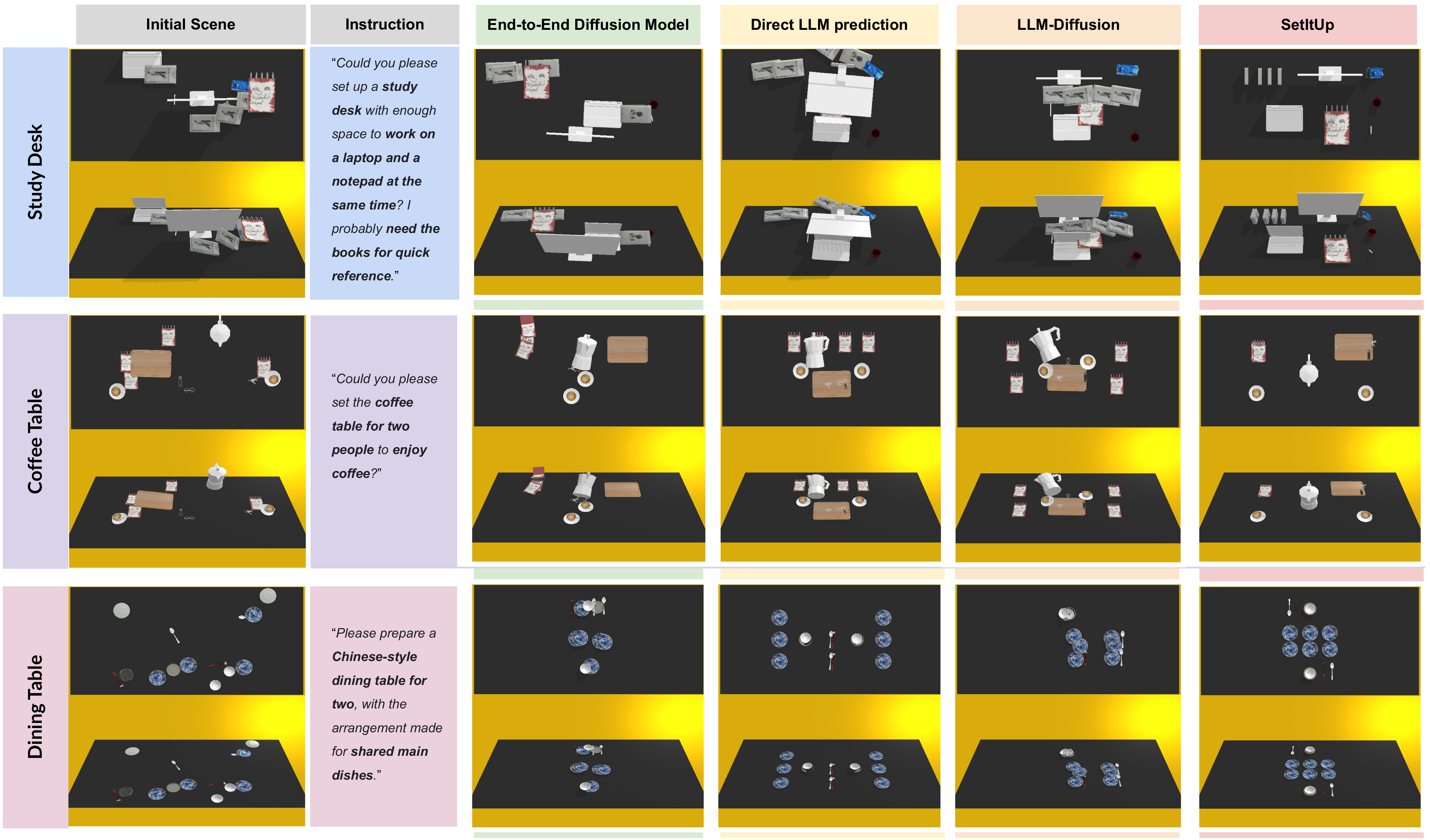}
    \caption{Illustrations of the final arrangements generated by our method and the baselines. Our model consistently generates more physically plausible, functional, and aesthetically pleasing object arrangements.}
    \label{fig:visual-comparison}
\end{figure*}

%% file: fig-texts/in_vs_out_dist.tex
\begin{figure}[tp]
    \centering
    \includegraphics[width=\linewidth]{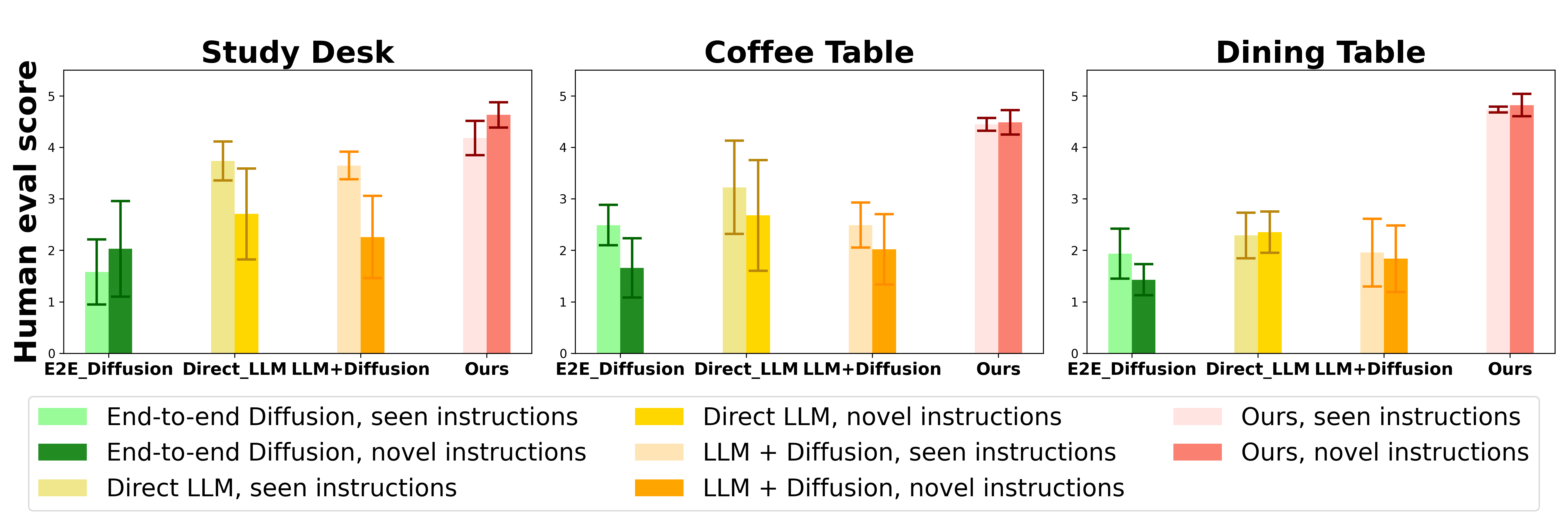}
    \caption{Evaluation on the generalization to novel instructions. Details on the seen-unseen splits are provided in the appendix. Our model shows the least amount of performance drop when generalizing to novel instructions.}
    \label{fig:in_vs_out_dist}
\end{figure}

%% file: fig-texts/scalability.tex
\begin{figure}[tp]
    \centering
    \includegraphics[width=\linewidth]{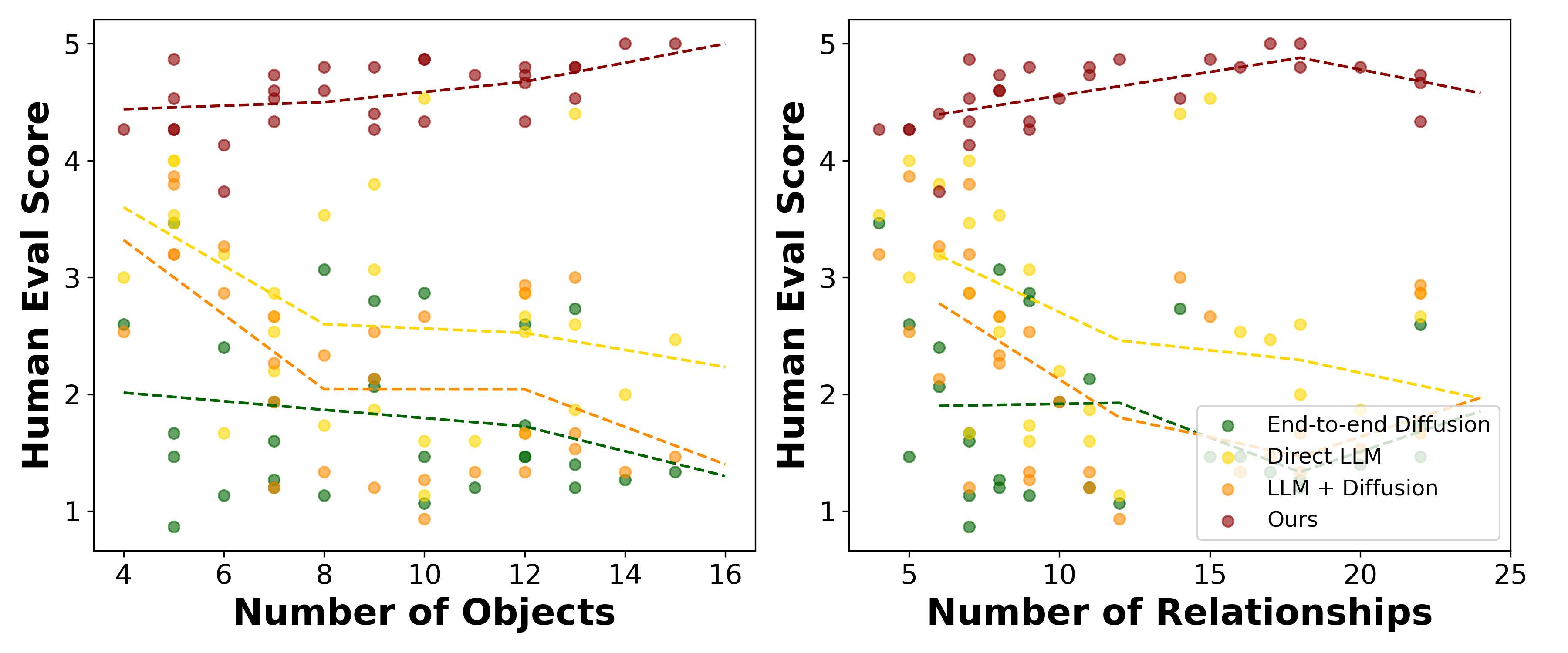}
    \caption{Evaluation of the model performance on scenes with different numbers of objects and relationships. The dashed lines depict the mean human evaluation scores for each method. As the number of objects and active relationships increases, our model consistently produces layouts that satisfy human evaluators, while all baselines have noticeable performance declines.}
    \label{fig:scalability}
\end{figure}

%% file: sections/2-related.tex
\section{Related Work}

\subsection{Object Rearrangement}
The literature on robotic object rearrangement is extensive, with many studies assuming that a goal arrangement is provided, detailing the precise object positions ~\citep{danielczuk2021object, driess2021learning, goodwin2022semantically, manuelli2019kpam, qureshi2021nerp, simeonov2021long, simeonov2023se, yuan2022sornet, zeng2021transporter}. In many cases, they focus on the detailed task and motion planning (TAMP), which comes with well-defined locations or using goal images as references. Another class of works involves interpreting language instructions to determine object placement by specified relationships \cite{gkanatsios2023energy}. ALFRED introduced a multi-step, language-guided rearrangement task ~\cite{blukis2022persistent, shridhar2020alfred}, inspiring solutions that merge high-level skills. However, these methods function at a basic level, focusing solely on placing objects in the correct receptacles without considering the necessary functional and spatial relationships among objects for effective tabletop arrangements. Our method differs from past approaches that require detailed goal specifications or concrete language instructions. Instead, we work with under-specified human instructions, inferring the necessary abstract relationships between objects within a scenario and determining their precise arrangements to fulfill them.

\subsection{Functional Object Arrangement}

Organizing objects on a tabletop to reach neatly functional arrangements is distinct from mere geometric rearrangement. It requires that objects not only appear orderly but also fulfill their intended purpose, often involving arranging like-function objects together.

Several benchmarks  \cite{szot2021habitat, weihs2021visual, kant2022housekeep}  address room tidying and common-sense notions of neatness at a basic level, focusing on correctly placing items in designated containers to satisfy semantic and aesthetic criteria \cite{wu2023tidybot, sarch2022tidee}. However, these do not account for the spatial relationships among the objects crucial for both functional and visual neatness on a tabletop. Previous data-driven research, such as Structformer \cite{liu2022structformer}  and StructDiffusion\cite{liu2022structdiffusion} focused on predicting object arrangements from vague instructions. They trained a single transformer and a diffusion model respectively to propose the placements of the objects. Other approaches trained a single model to predict the tidiness score \cite{kapelyukh2022my, wei2023lego} or the gradient towards an orderly configuration \cite{wu2022targf, kapelyukh2022my}. These methods often rely on extensive datasets and struggle to accommodate new object combinations or specific user preferences. Advancing towards a more generalizable object arrangement proposal, DALL-E-BOT \cite{kapelyukh2023dall} employs a pre-trained VLM to propose commonsense arrangements for open-category objects. However, its sole reliance on a single VLM for zero-shot layout generation from object types prevents it from accommodating user preferences and intended functional arrangements. Moreover, it overlooks object geometries, essential for creating physically feasible arrangements.

Our work differs by enabling zero-shot performance with novel items in various contexts through our symbolic reasoning powered by an LLM and compositional diffusion model grounding, eliminating the need for expert demonstrations.

\subsection{Knowledge Extraction from LLM}
Recent advancements have led to the development of various large language models (LLMs) capable of encoding extensive common sense knowledge and aiding in robot decision-making tasks \cite{liu2023pre}. These models have shown promise in domains such as task planning for household robotics  \cite{ahn2022can, huang2022language,  wu2023tidybot}. Huang et al. demonstrated that iterative prompt augmentation with LLMs improves task plan generation \cite{huang2022language}. Meanwhile, the SayCan approach integrates affordance functions into robot planning, facilitating action feasibility assessment when processing natural language service requests like ``make breakfast" \cite{ahn2022can}. LLM-GROP uses LLM to suggest object placements through prompts \cite{ding2023task}

Despite these advancements, LLMs often fails to capture the spatial understanding when organizing spaces, especially as object count and environmental complexity increase. Some research has tried to mitigate this issue by suggesting object placements using LLMs and then grounding these suggestions using vision-based models trained on orderly configurations \cite{xu2023tidy}. However, this method typically underperforms with novel object arrangements due to its reliance on a single visual model. Our approach differs markedly, as we employ compositional diffusion models to anchor symbolic object relationships accurately. The inherent compositional capabilities of these models offer a principled method for optimizing relational sets, enhancing our ability to handle complex geometric arrangements effectively.

%% file: sections/conclusion.tex
\section{Conclusion}

We have proposed \model, a neuro-symbolic model for compositional commonsense object arrangements. In order to achieve strong data efficiency and generalization, the design of \model is based on two important principles. First, we leverage large language models as a commonsense knowledge base to generate arrangement plans in an abstract language, based on simple geometric relationships. Second, in order to find global scene arrangements satisfying all proposed relationships, we use a compositional diffusion model as a continuous constraint satisfaction problem solver. We show that by composing individually trained diffusion models on synthetic data at test time, our system directly generalizes scenes with many objects and many relationships. This algorithm addresses the data scarcity issue of large scene arrangements and has strong extensibility. Both rule-based and human evaluations show that our model is capable of generating more physically feasible, functional, and aesthetic object placements compared to both pure LLM-based and end-to-end neural generative model baselines.

%% file: sections/appendix.tex
\section{Appendix} 
\label{sec:appendix}

\input{appendix/01-sampling}
\input{appendix/02-prompt}
\input{appendix/03-abstract_rule_classifier}

\input{appendix/04-diffusion_model}
\input{appendix/05-baselines}
\input{appendix/06-task_family}

\input{appendix/07-robot_setup}
\input{appendix/08-adding_new_relationship}
\input{appendix/09-ablation_study}
\input{appendix/10-self-reflection}
\input{appendix/11-additional_stats_test}
\input{appendix/12-user_generated_sketch}

%% file: appendix/01-sampling.tex
\subsection{Reverse Sampling and ULA Sampling}
\label{sect:reverse_ula}

In this section, we illustrate how a step of reverse sampling at a fixed noise level in a diffusion model is equivalent to ULA sampling at the same fixed noise level. The reverse sampling step on an input $x_t$  at a fixed noise level at timestep $t$ is given by by a Gaussian with a mean 
\begin{equation*}
        \mu_\theta(x_t, t) = x_t - \frac{\beta_t}{\sqrt{1 - \bar{\alpha}_t}}\epsilon_\theta(x_t, t).
\end{equation*}
with the variance of $\beta_t$ (using the variance small noise schedule in~\cite{ho2020denoising}). This corresponds to a sampling update,
\begin{equation*}
        x_{t+1} = x_t - \frac{\beta_t}{\sqrt{1 - \bar{\alpha}_t}}\epsilon_\theta(x_t, t) + \beta_t \xi, \quad \xi \sim \mathcal{N}(0, 1).
\end{equation*}

Note that the expression $\frac{\epsilon_\theta(x_t, t)}{\sqrt{1 - \bar{\alpha}_t}}$ corresponds to the score $\nabla_x p_t(x)$, through the denoising score matching objective~\citep{vincent2011connection}, where $p_t(x)$ corresponds to the data distribution perturbed with $t$ steps of noise. The reverse sampling step can be equivalently written as
\begin{equation}
        \label{eqn:reverse2}
        x_{t+1} = x_t - \beta_t \nabla_x p_t(x) + \beta_t \xi, \quad \xi \sim \mathcal{N}(0, 1).
\end{equation}

The ULA sampler draws a MCMC sample from the probability distribution $p_t(x)$ using the expression
\begin{equation}
        \label{eqn:ula}
        x_{t+1} = x_t - \eta \nabla_x p_t(x) + \sqrt{2} \eta \xi, \quad \xi \sim \mathcal{N}(0, 1),
\end{equation}
where $\eta$ is the step size of sampling.

By substituting $\eta=\beta_t$ in the ULA sampler, the sampler becomes
\begin{equation}
        \label{eqn:ula_beta}
        x_{t+1} = x_t - \beta_t \nabla_x p_t(x) + \sqrt{2} \beta_t \xi, \quad \xi \sim \mathcal{N}(0, 1).
\end{equation}
Note that the sampling expression for ULA sampling in Eqn~\ref{eqn:ula_beta} is very similar to the sampling procedure in the standard diffusion reverse process in Eqn~\ref{eqn:reverse2}, where there is a factor of $\sqrt{2}$ scaling in the ULA sampling procedures. Thus, we can implement ULA sampling equivalently by running the standard reverse process, but by scaling the variance of the small noise schedule by a factor of 2 (essentially multiplying the noise added in each timestep by a factor of $\sqrt{2}$). Alternatively, if we directly use the variance of the small noise schedule, we are running ULA on a tempered variant of $p_t(x)$ with temperature $\frac{1}{\sqrt{2}}$ (corresponding to less stochastic samples).

%% file: appendix/02-prompt.tex
\subsection{Prompts for \model}
\label{ssec:prompt}

Our query prompt for the LLM to identify abstract spatial relationships is composed of four parts:
\begin{enumerate}
    \item a task prompt including the relationship library,
    \item few-shot examples, 
    \item a human-crafted program sketch for each task family,
    \item a self-reflective and chain-of-thought (CoT) output prompt for querying relationships in a new task instance.
\end{enumerate}

The initial three steps focus on creating a well-documented, partially instantiated rule template for each task family. The final step involves inputting new test instance information to generate corresponding object relationship proposals. Below, we outline the specific prompts used in each section and provide the program sketches for each task family.

\subsubsection{Task Prompt with Spatial Relationship Library}

We use the same task prompt across all task families to guide the LLM in 1) proposing object relationships for \taskname object arrangement (FORM), and 2) presenting all available relationships from the relationship family $\mathcal{R}$. Crucially, each relationship is named to reflect its geometric property and includes a description to assist the LLM in understanding each relationship, facilitating the composition of the final constraint graph.

\begin{lstlisting}
    I would like to create an arrangement of objects that is both orderly (physically feasible and visually appealing) and functional (meets the user's intended purpose as per instruction), referred to as FORM, based on an under-specified user instruction. I need assistance in developing a Python rule template for generating these functional object arrangements for <TASK FAMILY> using only the abstract object relationships 

    We provide the following library of abstract relationships:
        
        - near_front_edge(Obj_A): Unary. Obj_A is positioned near the table's front edge, appearing at the bottom in the top-down camera view and closest in the front view. The shortest distance from any point on the bounding box to the front edge is below a specified threshold.

        - near_back_edge(Obj_A): Unary. Obj_A is near the table's back edge (the edge opposite the front). The shortest distance from any point on the bounding box to the back edge is below a specified threshold.

        - near_left_edge(Obj_A): Unary. Obj_A is near the table's left edge, on the left in both top-down and front views. The shortest distance from any point on the bounding box to the left edge is below a specified threshold.
        
        - near_right_edge(Obj_A): Unary. Obj_A is near the table's right edge, on the right in both top-down and front views. The shortest distance from any point on the bounding box to the right edge is below a specified threshold.
        
        - front_half(Obj_A): Unary. Obj_A entirely resides within the table's front half, occupying the front portion as divided along the table's horizontal center axis. The highest point of the bounding box is below the table's horizontal centerline.
        
        - back_half(Obj_A): Unary. Obj_A entirely resides within the table's back half, occupying the back portion as divided along the horizontal center axis. The lowest point of the bounding box is above the table's horizontal centerline.
        
        - left_half(Obj_A): Unary. Obj_A entirely resides within the table's left half, dividing it vertically along the center axis. The rightmost point of the bounding box is on the left of the table's vertical centerline.
        
        - right_half(Obj_A): Unary. Obj_A entirely resides within the table's right half, divided vertically along the center axis. The leftmost point of the bounding box is on the right of the table's vertical centerline.
        
        - central_column(Obj_A): Unary. The centroid of Obj_A is within the central column of the table, which runs vertically through the table's midpoint, covering half the table's width. The centroid_A is within the central column bounds.
        
        - central_row(Obj_A): Unary. The centroid of Obj_A is within the central row of the table, running horizontally through the midpoint and covering half the table's length. The centroid_A is within the central row bounds.
        
        - centered_table(Obj_A): Unary. The centroid of Obj_A coincides with the center of the table. The distance between centroid_A and the table's center is below a threshold.
        
        - horizontally_aligned(Obj_A, Obj_B): Binary. Obj_A and Obj_B are horizontally aligned at their bases. The differences in bottom coordinates and angles of their bounding boxes are below a threshold.
        
        - vertically_aligned(Obj_A, Obj_B): Binary. Obj_A and Obj_B are vertically aligned at their centroids. The vertical distance between their centroids and the angles of their bounding boxes are below a threshold.
        
        - left_of(Obj_A, Obj_B): Binary. Obj_A is immediately to the left of Obj_B, viewed from the canonical camera frame. The right side of Obj_A and the left side of Obj_B are within a threshold, with substantial y-overlap.
        
        - right_of(Obj_A, Obj_B): Binary. Obj_A is immediately to the right of Obj_B, viewed from the canonical camera frame. The left side of Obj_A and the right side of Obj_B are within a threshold, with substantial y-overlap.
        
        - on_top_of(Obj_A, Obj_B): Binary. Obj_A is placed atop Obj_B. Obj_A's bounding box completely overlaps with Obj_B's, and Obj_B's area is equal or larger.
        
        - centered(Obj_A, Obj_B): Binary. The center of Obj_A coincides with that of Obj_B. The distance between their centroids y-coordinates is below a threshold.
        
        - vertical_symmetry_on_table(Obj_A, Obj_B): Binary. Obj_A and Obj_B mirror each other across the table's vertical axis. The centroids are mirrored along the vertical centerline.
        
        - horizontal_symmetry_on_table(Obj_A, Obj_B): Binary. Obj_A and Obj_B mirror each other across the table's horizontal axis. The centroids are mirrored along the horizontal centerline.
        
        - vertical_line_symmetry(Axis_obj, Obj_A, Obj_B): Ternary. Obj_A and Obj_B are symmetrical with respect to Axis_obj's vertical axis. Mirrored centroids along Axis_obj's vertical line.
        
        - horizontal_line_symmetry(Axis_obj, Obj_A, Obj_B): Ternary. Obj_A and Obj_B are symmetrical with respect to Axis_obj's horizontal axis. Mirrored centroids along Axis_obj's horizontal line.
        
        - aligned_in_horizontal_line(Obj_1, ..., Obj_N): Variable (N). Objects are horizontally aligned with equal spacing. Pairwise bottom coordinate differences and angle variations of bounding boxes are below a threshold, ensuring equal horizontal spacing.
        
        - aligned_in_vertical_line(Obj_1, ..., Obj_N): Variable (N). Objects are vertically aligned with equal spacing. Pairwise centroid y-coordinates' differences and bounding box angle variations are below a threshold, ensuring equal vertical spacing.
        
        - regular_grid(Obj_1, ..., Obj_N): Variable (N). Objects are arranged in a grid or matrix pattern. Centroids in the same row or column are aligned, with constant distances between rows and columns.

The functional object arrangement for a <TASK FAMILY> should be grounded strictly using these abstract object relationships.

Your task is to create a rule template for FORM (Functional Object Arrangement) applicable to <TASK FAMILY> in Python syntax. This template will summarize and abstract the common rules and patterns necessary for FORM generation. You will be provided with 5 examples of functional arrangements for <TASK FAMILY> alongside a sketch of a human-designed program. Please ensure you receive all examples and the program sketch before you start crafting the rule template.

\end{lstlisting}

\subsubsection{Few-shot Examples}
For each task family, we input five examples of the FORM arrangement. Each example includes i) a user instruction outlining desired object arrangements, ii) the list of objects present on the table, and iii) the active abstract relationships. Objects are referred to by their types, allowing the LLM to deduce their functions. Abstract relationships are annotated by the abstract rule classifiers, as detailed in Appendix \ref{apped:abstract_rule}, based on objects' shapes and poses.

\begin{lstlisting}
I will provide 5 examples of an organized <TASK FAMILY> sequentially. Please wait to receive all examples and the program sketch before creating the rule template.

Example 1:

- User instruction: Please arrange a study desk for my work, primarily with my laptop, and I'll need the books for quick reference too.

- Object list: ["laptop", "book_1", "book_2", "book_3", "book_4", "mug"]

- Active Relationships: (Identified using abstract rule classifiers)

----
More examples
...

Ensure you have all examples and the program sketch before beginning the template creation.
\end{lstlisting}

\subsubsection{Task-family Program Sketch}

The task-family program sketch is a concise program designed by a human to decompose arrangement tasks into simpler sub-problems, each with simpler input and output criteria for easy instantiation by the LLM. The program sketch, coded in Python, generally starts by extracting task-relevant data from user instructions and the object list. Then, it sorts objects into functional groups and identifies the active relationships within each group, as well as between groups.

Here is the program sketch for organizing the study table layout.
\begin{lstlisting}[language=Python]
    def study_desk_layout(instruction, object_list):
        """
        Setting up the study table based on the user instructions and a list of objects.
            
        Workflow:
        1. Extract the main devices for the study activities.
        2. Categorize the main devices into input, output, and IO devices. 
        3. For each category, find their associated objects and determine the placements for both the main devices and their associated objects.
            
        Parameters:
        - user_instruction (str): Instruction for setting up, highlighting any special preferences.
        - object_list (list): Available objects for arrangement, denoted by their names.
            
        Returns:
        - List of active relationships delineating the arrangement of objects.
        """
         # Infer the main purpose-driven devices
         main_devices = extract_main_devices(instruction, object_list)
        
         # Categorize devices based on their function
         input_devices, output_devices, IO_devices = categorize_main_devices_based_on_functions(main_devices)
    
         active_relationships = []
        
         # Input Devices Placement
         # Associate input devices with their related objects
         associated_objects = associating_objects(input_devices, object_list)
         # For input devices, place them near the front edge
         active_relationships = locate_input_devices_and_their_associates(input_devices, associated_objects, active_relationships)
        
         # Output Devices Placement
         # Associate output devices with their related objects
         associated_objects = associating_objects(output_devices, object_list)
         # For output devices, place them near the back edge
         active_relationships = locate_output_devices_and_their_associates(output_devices, associated_objects, active_relationships)
                   
         # Input and Output Devices Placement
         # Associate IO_devices with their related objects
         associated_objects = associating_objects(IO_devices, object_list)
         # For IO_devices, placed them at the center
         active_relationships = locate_IO_devices_and_their_associates(IO_devices, associated_objects, active_relationships)
          
         # Remaining Objects Placement
         active_relationships = locate_the_remaining_objects_based_on_commonsense(object_list, active_relationships) 

         return active_relationships
\end{lstlisting}

Here is the program sketch for organizing the coffee table layout.
\begin{lstlisting}[language=Python]
    def coffee_table_layout(user_instruction, object_list):
        """
        Arranges objects on a coffee table based on user instructions and intended activities.
        
        The function considers whether the table will be used for specific activities or for general storage and decoration. It then organizes objects to support these uses, focusing on ease of retrieval, aesthetics, and functional needs.
        
        Parameters:
        - user_instruction (str): Instructions or description of desired table use provided by the user.
        - object_list (list): List of objects available for arrangement.
        
        Returns:
        - List of tuples describing the active relationships among objects, indicating their arrangement.
        """
    
        # Determine the main activity to guide the arrangement. If there is no activity requested, the default is "storage&decoration".
        main_activity = infer_main_activities(user_instruction)
    
        # Extract objects designated for storage or decoration.
        storage_objects, decoration_objects = extract_storage_and_decoration_objects(object_list)
    
        # Initialize list for active object relationships.
        active_relationships = []
    
        if main_activity == "storage&decoration":  # Default layout for storage and aesthetic purpose.
    
        	  # Placement of the storage objects.
            # Place storage objects for easy access, typically near the table's front edge.
            active_relationships = locate_storage_objects_as_main_activities(storage_objects, active_relationships)
    
            # Placement of the decoration objects.
            # Arrange decoration objects centrally with an emphasis on symmetry and visual appeal.
            active_relationships = locate_decoration_objects_as_main_activities(decoration_objects, active_relationships)
    
        else:  # Specific activities identified, necessitating a functional arrangement.
    
            # Estimate the number of participants involved in the activity, which can range from 1 to N, where N signifies a group activity with a variable number of participants.
            num_of_participants = infer_the_number_of_participants(user_instruction)
    
            # Determine participants' locations around the table.
            participants_locations = infer_participants_locations(user_instruction, num_of_participants)
    
            # Categorize remaining objects by their relation to the activity: shared, individual, or unrelated.
            shared_objects, individual_objects, remaining_objects = categorize_objects_by_usage(object_list, storage_objects, decoration_objects)
    
            # Placement of the shared objects.
            # Evenly space shared objects at the table center for communal access.
            active_relationships = locate_shared_objects(shared_objects, active_relationships)

            # Placement of the individual objects by:
        		# 1. Assigning specific objects to each participant.
        		# 2. Establishing each object's position relative to the table, considering participant locations.
        		# 3. Organizing the layout of objects in each participant's designated area for optimal use and accessibility.
                active_relationships = locate_individual_objects(individual_objects, num_of_participants, participants_locations, active_relationships)
    
            # Placement of the storage objects.
            # Store storage objects towards the table's rear to avoid disrupting the primary activity.
            active_relationships = locate_storage_objects(storage_objects, active_relationships)
    
            # Placement of the decoration objects.
            # Place decoration objects in the table's back half within the central column for aesthetic without hindrance.
            active_relationships = locate_decoration_objects(decoration_objects, active_relationships)
    
            # Placement of the remaining objects.
            # Optimize the placement of remaining objects to minimize space usage and maintain order.
            active_relationships = locate_remaining_objects(remaining_objects, active_relationships)
    
        return active_relationships
\end{lstlisting}

Here is the program sketch for organizing the dining table layout.

\begin{lstlisting}[language=Python]
    def dining_table_layout(user_instruction, object_list):
        """
        Organizes the dining table layout influenced by user instructions and an inventory of objects.
        
        Workflow:
        1. Determines the number of diners.
        2. Identifies seating positions for the diners.
        3. Categorizes objects into shared dishes, individual cutleries, and miscellaneous items.
        4. Positions the shared dishes and individual cutleries according to dining etiquette.
        5. Aligns miscellaneous items neatly around the setup.
        
        Parameters:
        - user_instruction (str): Instruction for setting up, highlighting any special preferences.
        - object_list (list): Available objects for arrangement, denoted by their names.
        
        Returns:
        - List of active relationships delineating the arrangement of objects.
        """

        # Infer the number of diners from user instructions.
        number_of_diners = infer_the_number_of_diners(user_instruction)

        # Deduce seating positions considering the number of diners and their preferences.
        # By default, the diner sits facing the edge closest to the camera (front edge). The next diner facing them will sit facing the table's back edge. 
        # Example: diners_seating_positions[diner_0] = [central_column, near_front_edge]
        # If diners need to sit on the same side, they'll start by the front edge, one on the right and the other on the left.
        # Example: diners_seating_positions[diner_0] = [left_half, near_front_edge]
        # Ensure diners are spaced evenly.
        diners_seating_positions = infer_the_diners_seating_positions(user_instruction, number_of_diners)

        # Categorize objects. Shared dishes are for communal use, individual cutleries for personal use, and 'others' for non-dining items like decor.
        shared_dishes, individual_cutleries, others = categorize_objects_based_on_ownership(user_instruction, number_of_diners, object_list)

        active_relationships = []  # Holds the spatial relationships between objects.

        # Position shared dishes at the table's center, ensuring even distribution.
        # Example: Align shared dishes in a central row or in a regular grid for multiple items.
        active_relationships = locate_shared_dishes(shared_dishes, active_relationships)

        # Arrange individual cutleries per diner, tailored to conventional dining setups.
        # Associate each cutlery set with a diner, placing items based on diner positions and observing dining etiquette.
        # The LLM leverages common sense to adapt the arrangement to the style of dining specified. 
        # Note: The terms "left_of" and "right_of" are from the perspective of a diner seated facing the front edge of the table.
        # For diners seated opposite, facing the back edge, these directions are inverted to match their point of view.
        active_relationships = locate_individual_cutleries(individual_cutleries, number_of_diners, diners_seating_positions, active_relationships)

        # Strategically place remaining items (e.g., seasoning bottles) to ensure accessibility without clutter.
        # Seek alignment or aesthetic arrangement, like organizing them in a neat row along one edge.
        active_relationships = locate_the_remaining_objects(others, active_relationships)
        
        # Factor in arrangements that consider the relationships between different groups of objects, enhancing overall table symmetry and balance.
        # Example: Ensure serving plates across from each other are symmetrically positioned for diners facing one another.
        active_relationships = add_inter_group_relationships(shared_dishes, individual_cutleries, others, active_relationships)

        return active_relationships
\end{lstlisting}

Finally, we prompt LLM to generate the task-family FORM rule template based on the program sketch and a few examples.

\begin{lstlisting}
    The generated rule template for <TASK FAMILY> should be based on the program sketch below. It must 1) include detailed documentation for each function, outlining its input, output, intended purpose, and examples, and 2) use the abstract relationships to partially instantiate the functions by identifying common arrangement patterns from the provided examples.
\end{lstlisting}

\subsubsection{Self-reflective \& CoT Output Prompt}
Given a new task instance, we use the following prompt to query the abstract relationship set for the new setting. 
\begin{lstlisting}
    Great work! For the next step, utilize the rule template to produce an <TASK FAMILY> setup with a given group of objects. Follow these steps for your output:

    1. Utilizing the generated program as a guide, devise a natural language description for a dining table arrangement.
    2. Fully instantiate the rule template based on the given instruction and the object list, grounding the FORM arrangement using the 24 defined abstract relationships.
    3. Given the proposed relation set, create a natural language-based description of the arrangement.
    4. Verify coherence between the descriptions in steps 1 and 3. If inconsistency found, repeat step 1-4.

    Take note that the "left_of/right_of" relation for the diner seating at the back edge is reversed. For instance, instead of 'right_of(knife, plate)' for placing the knife on the right of the back-edge diner, it should be 'left_of(knife, plate)'.

    During step 2, the relations should be sequenced as follows:
    
    2.1. Begin by listing all objects.
        
    2.2. For each listed object:
       2.2.1. Determine its position about the table.
       2.2.2. Ascertain its relation to other objects.
    
    Here is the new user instruction and the object list:
    User instruction: [Insert user instruction here]
    Object list: [Insert object list here]
    
    Your task is to arrange these objects using the 24 types of abstract relationships to create an FORM (Orderly Functional Arrangement). Extract all proposed relationships into a list, following this format:
       [["near_front_edge", "serving_plate_1"],
       ["near_back_edge", "serving_plate_2"],
       ...]
\end{lstlisting}

%% file: appendix/03-abstract_rule_classifier.tex
\subsection{Spatial Relationship Library}
\label{apped:abstract_rule}
Our spatial relationship library consists of the following 24 relationships.
We use these relationships as the intermediate representation for generating the abstract object arrangements using the LLM and grounding these abstract spatial relationships using the compositional diffusion models. In the following section, we describe the geometric concepts of these relationships and the implementation of their classifiers $h_R$.

\begin{longtable}[ht]{>{\raggedright\arraybackslash}p{3.5cm}>{\raggedright\arraybackslash}p{2cm}>{\raggedright\arraybackslash}p{4.5cm}>{\raggedright\arraybackslash}p{4.5cm}}

\toprule
\textbf{Relationship} & \textbf{Arity} & \textbf{Description} & \textbf{Implementation of $h_R$} \\
\midrule
near\_front\_edge(Obj\_A) & Unary & Obj\_A is positioned near the table's front edge, which appears at the bottom in the top-down camera view and closest to the camera in the front view. & The shortest distance from any point on the bounding box to the front edge falls below a specified threshold. \\
\midrule
near\_back\_edge(Obj\_A) & Unary &  Obj\_A is near the back edge of the table (i.e. the edge opposite the front edge).  & The shortest distance from any point on the bounding box to the back edge falls below a specified threshold.\\
\midrule
near\_left\_edge(Obj\_A) & Unary & Obj\_A is near the left edge of the table, which is on the left in both top-down and front views. & The shortest distance from any point on the bounding box to the left edge falls below a specified threshold. \\
\midrule
near\_right\_edge(Obj\_A) & Unary & Obj\_A is near the right edge of the table, which is on the right in both top-down and front views. & The shortest distance from any point on the bounding box to the right edge falls below a specified threshold. \\
\midrule
front\_half(Obj\_A) & Unary & Obj\_A entirely resides within the front half of the table, occupying the front portion as divided horizontally along the table's center axis. & Check if the highest point of the bounding box is below the table's horizontal centerline. \\
\midrule
back\_half(Obj\_A) & Unary & Obj\_A entirely resides within the back half of the table, occupying the back portion as divided horizontally along the table's center axis. & Check if the lowest point of the bounding box is above the table's horizontal centerline. \\
\midrule
left\_half(Obj\_A) & Unary & Obj\_A entirely resides within the left half of the table, occupying the left portion as divided vertically along the table's center axis. & Check if the rightmost point of the bounding box is on the left of the table's vertical centerline. \\
\midrule
right\_half(Obj\_A) & Unary & Obj\_A entirely resides within the left half of the table, occupying the right portion as divided vertically along the table's center axis. & Check if the leftmost point of the bounding box is on the right of the table's vertical centerline.  \\
\midrule
central\_column(Obj\_A) & Unary & The centroid of obj\_A is positioned within the central column of the table, which runs vertically through the table's midpoint, spanning half the table's width. & Check if centroid\_A is within the central column bounds. \\
\midrule
central\_row(Obj\_A) & Unary & The centroid of obj\_A is positioned within the central row of the table, which runs horizontally through the table's midpoint, spanning half the table's length. & Check if centroid\_A is within the central row bounds. \\
\midrule
centered\_table(Obj\_A) & Unary & The centroid of Obj\_A coincides with center of the table. & Check if the distance between the centroid\_A and the table's center is below a threshold. \\
\midrule
horizontally\_aligned (Obj\_A, Obj\_B) & Binary & Obj\_A and Obj\_B are aligned horizontally at their bases. & Check if $|\textit{BB}_A.\textit{bottom} - \textit{BB}_B.\textit{bottom}|< \textit{threshold}$ and  $|\textit{BB}_A.\theta - \textit{BB}_B.\theta|< \textit{threshold}$ \\
\midrule
vertically\_aligned (Obj\_A, Obj\_B) & Binary & Obj\_A and Obj\_B are aligned vertically at their centroids. & Check if $|\textit{centroid}_A.y - \textit{centroid}_B.y| < \textit{threshold}$ and  $|\textit{BB}_A.\theta - \textit{BB}_B.\theta| < \textit{threshold}$ \\
\midrule
left\_of(Obj\_A, Obj\_B) & Binary & Obj\_A is immediately to the left side of Obj\_B viewed from the canonical camera frame. & Check if 1) $|\textit{BB}_A.\textit{right} - \textit{BB}_B.\textit{left}| < \textit{threshold}$, and 2) $\textit{BB}_A.y$ and $\textit{BB}_B.y$ overlap substantially.\\
\midrule
right\_of(Obj\_A, Obj\_B) & Binary & Obj\_A is immediately to the right side of Obj\_B viewed from the canonical camera frame. & Check if 1) $|\textit{BB}_A.\textit{left} - \textit{BB}_B.\textit{right}| < \textit{threshold}$, and 2) $\textit{BB}_A.y$ and $\textit{BB}_B.y$ overlap substantially.\\
\midrule
on\_top\_of(Obj\_A, Obj\_B) & Binary & Obj\_A is placed on top of Obj\_B. & Check if 1) $\textit{BB}_A$ completely overlaps with $\textit{BB}_B$, and 2) the area of $\textit{BB}_B$ is equal or larger than $\textit{BB}_A$.\\
\midrule
centered(Obj\_A, Obj\_B) & Binary & Center of Obj\_A coincides with center of Obj\_B. & Check if $|\textit{centroid}_A.y - \textit{centroid}_B.y| < \textit{threshold}$. \\
\midrule
vertical\_symmetry\_on
\_table(Obj\_A, Obj\_B) & binary & Obj\_A and Obj\_B are placed on a table mirroring each other across the vertical axis. & Check mirrored centroids along the vertical centerline of the table. \\
\midrule
horizontal\_symmetry\_on
\_table(Obj\_A, Obj\_B) & Binary & Obj\_A and Obj\_B are placed on a table mirroring each other across the horizontal axis. & Check mirrored centroids along the horizontal centerline of the table. \\
\midrule
vertical\_line\_symmetry
(Axis\_obj, Obj\_A, Obj\_B) & Ternary & Obj\_A and Obj\_B are symmetrical to each other with respect to a vertical axis of the Axis\_obj, meaning they are mirror images along that axis. & Check mirrored centroids along Axis\_obj's vertical line \\
\midrule
horizontal\_line\_symmetry
(Axis\_obj, Obj\_A, Obj\_B)  & Ternary & Obj\_A and Obj\_B are symmetrical to each other with respect to the horizontal axis of the Axis\_obj, meaning they are mirror images along that axis. & Check mirrored centroids along Axis\_obj's horizontal line \\
\midrule
aligned\_in\_horizontal
\_line(Obj\_1, ..., Obj\_N) & Variable (N) & Objects aligned horizontally with equal spacing. & Check for pairwise $|\textit{BB}_i.\textit{bottom} - \textit{BB}_j.\textit{bottom}|< \textit{threshold}$ and  $|\textit{BB}_i.\theta - \textit{BB}_j.\theta|< \textit{threshold}$, and equal spacing horizontally between the adjacent objects. \\
\midrule
aligned\_in\_vertical
\_line(Obj\_1, ..., Obj\_N) & Variable (N) & Objects aligned vertically with equal spacing. & Check for pairwise $|\textit{centroid}_i.\textit{y} - \textit{centroid}_j.\textit{y}|< \textit{threshold}$ and  $|\textit{BB}_i.\theta - \textit{BB}_j.\theta|< \textit{threshold}$, and equal spacing vertically between the adjacent objects. \\
\midrule
regular\_grid(Obj\_1, ..., Obj\_N) & Variable (N) & Arranged in a grid or matrix pattern. & "Verify if the centroids of objects in the same row or column are aligned and ensure the distances between rows and columns are constant.\\
\bottomrule
\label{tab:relationships}
\end{longtable}

%% file: appendix/04-diffusion_model.tex
\subsection{Diffusion Model Implementation Details}
\label{append:diffusion}

Each individual diffusion constraint solver is trained to predict the noise required for reconstructing object poses, thereby satisfying a specified spatial relationship. At its core, the model operates by taking normalized shapes and poses of objects, represented as 2D bounding boxes, and outputs the predicted noise for reconstructing these poses. The normalization process uses the dimensions of a reference table region to ensure consistent scaling.

\noindent
\textbf{Encoder Implementation}: The model features three primary encoders - a shape encoder, a pose encoder, and a time encoder, each tailored for its specific type of input. The shape and pose encoders leverage a two-layer neural network with SiLU activations, scaling the input dimensions to a predefined hidden dimension size 256. Specifically, the shape encoder transforms object geometry information, while the pose encoder processes the 2D bounding boxes' positions and sizes:
\begin{itemize}
    \item Shape Encoder: Processes geometry inputs using a sequential neural network architecture, first mapping from input dimensions to half the hidden dimension, followed by a mapping to the full hidden dimension.
    \item Pose Encoder and Pose Decoder: The pose encoder follows a structure similar to the shape encoder, focusing on the pose information of objects. Correspondingly, the pose decoder translates the hidden representation back to the pose dimensions, facilitating the reconstruction of object poses.
    \item Time Encoder: Comprises a sinusoidal positional embedding followed by a linear scaling to four times the hidden dimension, processed with a Mish activation, and then downscaled back to the hidden dimension.
\end{itemize}

\noindent
\textbf{Backbone Architecture}: Based on the arity of spatial relationships, our model employs two different backbones. A Multi-Layer Perceptron (MLP) serves as the backbone for relationships with fixed arity, utilizing a sequential architecture to process concatenated shape and pose features. For relationships of variable arity, a transformer backbone is employed, capable of handling varying sequence lengths and providing a flexible means to process an arbitrary number of objects:

\begin{itemize}
    \item MLP Backbone: Utilizes a linear layer followed by a SiLU activation, designed to intake normalized and concatenated object features for fixed arity relationships.
    \item Transformer Backbone: Processes variable-arity relationships by dynamically adjusting to the number of objects. The transformer is configured with a maximum capacity of 16 objects, determined by its sequence length limit of 16. It includes preprocessing steps for positional encodings, attention mask preparation, and padding to accommodate sequences of different lengths. The transformer's output is then post-processed and decoded into the pose dimensions.
\end{itemize}

\noindent
\textbf{Training and Inference}: The entire model, including encoders, backbone, and the pose decoder, is trained end-to-end with an L2 loss, aiming to minimize the difference between predicted and actual noise values used for pose reconstruction. During inference, our model utilizes a cosine beta schedule over 1500 diffusion steps to iteratively refine the noise predictions. For complex scenes with multiple spatial relationships, the model aggregates and averages (or weights) the noise features from relevant relationships before final decoding, enhancing pose prediction accuracy.

%% file: appendix/05-baselines.tex
\subsection{Baselines}
\label{append:baseline_implementation}

We have implemented three baseline models for our study: the End-to-end Diffusion Model, Direct LLM (Large Language Model) Prediction, and LLM + Diffusion (our ablation). Detailed descriptions of these models are provided below.

\vspace{1em}
\noindent
\textbf{End-to-end Diffusion Model:}
Drawing inspiration from StructDiffusion \cite{liu2022structdiffusion}, we developed a language-conditioned diffusion model. This model is designed to predict object poses in FORM arrangements. It has been trained using the same datasets as our main model. These datasets include a synthetic object relationship dataset and 15 specific FORM examples.

\vspace{0.5em}
For synthetic objects, we employ a script that describes the geometric relationships between scene objects as the language input. For FORM scenes, the language input comes from user instructions. We then encode these language inputs into language feature embedding using CLIP embedding. In our model, each object is represented by a combined token of its shape and pose embeddings. We have configured the transformer to handle up to 16 objects, which corresponds to its sequence length limit of 16. The model processes the language embeddings together with object sequence embeddings and time embeddings through a transformer encoder. This encoder uses four layers of residual attention blocks and is equipped with two heads. We also incorporate normalization layers both before and after the transformer layers. Finally, we utilize the same diffusion timesteps $t=1500$ in this model as in our \model.

\vspace{0.5em}
\textbf{Modification on StructDiffusion \cite{liu2022structdiffusion}}: Our adaptation maintains the core framework of StructDiffusion, which is a multimodal transformer that conditions on both the language embeddings from instructions and the objects' geometries. Key modifications are made for data format compatibility and architectural coherence.

\vspace{0.5em}

\begin{itemize}
    \item \textbf{Model Architecture}: We changed orginal 3D point cloud (Point Cloud Transformer) and a 6DOF pose encoder in StructDiffusion to the 2D geometry and pose encoder (same as \model) to fit the 2D bounding box object geometry and pose encoder used in \model. Both models utilize MLPs for diffusion encoding; our adaptation simply modifies the hidden layer parameters to align with the architecture of \model for consistent performance evaluation.
    \item \textbf{Different Training Data}: We use the training dataset for \model, which differs from the required training dataset for StructDiffusion. We changed from the orginal extensive dataset on broad paired instructions and scene-level object arrangements to the \model dataset, which only includes 1) 5 examples of the paired instructions and scene-level object arrangements, and 2) synthetic datasets labelled by their abstract relationships.
\end{itemize}

\vspace{0.5em}

These necessary modifications ensure that while the adapted StructDiffusion framework aligns closely with the original’s multi-modal transformative capabilities, it also enhances compatibility with \model’s data formats and model architecture choices. 

\vspace{1em}
\noindent
\textbf{Direct LLM prediction} 
Drawing on Tidybot's approach, we use an LLM-based method to directly propose the object poses for FORM arrangements. Specifically, we feed the LLM the same task instruction, along with a library of relationships and the few-shot examples. Additionally, we provide details of the table layout and include the object shapes in the few-shot training examples. It's important to note that, unlike our model, which only processes abstract relationships, the LLM is given actual shapes from the ground truth. We then instruct the LLM to identify common patterns within these examples, following the methodology proposed by Tidybot. For inference, we supply the user instructions, a list of objects, shapes of these objects to the LLM. Then, we ask the LLM to predict the poses for objects in FORM arrangements.

\vspace{0.5em}
Below, we detail the prompts used for few-shot learning and those used during inference.

\begin{lstlisting}[caption={Few-shot prompt for direct LLM-based prediction}]
    Description of the table: The table presents a rectangular shape with dimensions measuring 3 units in length and 2 units in width. The coordinate system is centered at the front-left corner of the table, with the origin (0, 0) located at the front-left corner of the tabletop. Consequently, the position of any object specified by the coordinates (x, y, \theta) must fall within the range: 0 < x < 3 along the length, 0 < y < 2 along the width, and \theta in radian.
  
    We have the following list of objects: laptop, book_1, book_2, book_3, book_4, mug.
    The objects have the following shapes in (l, w):
          laptop: (1.02, 0.60)
          book_1: (0.12, 0.40)
          book_2: (0.12, 0.40)
          book_3: (0.12, 0.40)
          book_4: (0.12, 0.40)
          mug: (0.25, 0.25)
    User instruction: Could you please set up a study desk for me? I will mainly be using it to work on my laptop. I probably need the books for quick reference.

    The objects are orderly and functionally arranged on the table. The following object relationships are active: [list of active relationships].
\end{lstlisting}

\begin{lstlisting}[caption={Output prompt for direct LLM-based prediction}]
    Task Description: The table presents a rectangular shape with dimensions measuring 3 units in length and 2 units in width. The coordinate system is centered at the front-left corner of the table, with the origin (0, 0) located at the front-left corner of the tabletop. Consequently, the position of any object specified by the coordinates (x, y, \theta) must fall within the range: 0 < x < 3 along the length, 0 < y < 2 along the width, and \theta in radian. Your task is to propose an functional object arrangement according to the user's instruction. 
    
    We have the following list of objects: laptop, mouse, lamp, mug, tissue_box.
    The objects have the following shapes in (l, w):
      laptop: (1.02, 0.60)
      mouse: (0.15, 0.20)
      lamp: (0.20, 0.30)
      mug: (0.25, 0.25)
      tissue_box: (0.25, 0.25)

    User Instruction: Could you please set up a study desk for me? I will mainly be using it to work on my laptop.
    To propose the tidy arrangements, strictly follow the following structure:
      - Object relationships that should be present: [List the tidy object relationships here]
      - The centroids and rotation of each object in a Python dictionary:
        object_name_1: [x_1, y_1, \theta_1]
        object_name_2: [x_2, y_2, \theta_2]
\end{lstlisting}

\vspace{0.5em}
\textbf{TidyBot Adaptation}: We use TidyBot's core strategy of using the summarization capabilities of the LLM to generalize from a limited number of examples to novel scenarios. Our adaptation adjust the summarization and pose inference mechanisms to better suit the specific needs of \model.

\vspace{0.5em}

\begin{itemize}
    \item \textbf{Summarization}: While TidyBot focuses on placing objects into designated receptacles, our objectives involve prescribing more complex and joint object poses. This necessitates richer and more intricate rules. To handle this complexity, we input additional information (so TidyBot-variant has the same information as \model), including the task instruction, sketch, abstract relationship library, and the CoT prompt.
    \item \textbf{Pose Inference}: Tidybot requires the LLM to output the abstract action primitives and the receptables for ``tidy arrangement", yet we need the object poses. Therefore, while the tasks in the Tidybot project requires no geometric understanding from LLM, FORM requires the LLM to be able to understand the geometries so as to propose the object poses. To help LLM with this, we provide the table layout and the object dimensions for few-shot learning.
\end{itemize}

\vspace{0.5em}

These adjustments ensure our adaptation of TidyBot not only retains the advantageous summarization for learning from minimal data but also meets the enhanced requirements of \model for precise and context-aware object placement.

\vspace{1em}
\noindent
\textbf{LLM + Diffusion:}

This model is ablation of our \model, specifically omitting the program sketch and the self-reflective \& Chain of Thought (CoT) output template while keeping the rest of the framework intact. This adjustment involves utilizing the same task prompt alongside the relationship library and employing a few-shot examples approach. Subsequently, we directly prompt the LLM to summarize the common arrangement rules and patterns in a natural language summary. At the inference stage, we supply the LLM with the user's instructions and the list of objects, requesting it to suggest potential active relationships. Below, we detail the prompt used during the inference process.

\begin{lstlisting}[caption={Output prompt for LLM + Diffusion Model}]
   Next, we use the common patterns summarized for <TASK FAMILY> to propose the active object relationships for a new user instruction and object list.
   
    Take note that the "left_of/right_of" relation for the diner seating at the back edge is reversed. For instance, instead of 'right_of(knife, plate)' for placing the knife on the right of the back-edge diner, it should be 'left_of(knife, plate)'.
    
    Here is the new user instruction and the object list:
    User instruction: [Insert user instruction here]
    Object list: [Insert object list here]
    
    Your task is to arrange these objects using the 24 types of abstract relationships to create an FORM ( Functional Object Arrangement). Extract all proposed relationships into a list, following this format:
       [["near_front_edge", "serving_plate_1"],
       ["near_back_edge", "serving_plate_2"],
       ...]
\end{lstlisting}

%% file: appendix/06-task_family.tex
\subsection{Task Family Overview}
\label{append:task_family}

This work encompasses three distinct task families: study desks, coffee tables, and dining tables. Each family represents a specific type of table setting, defined by its high-level purposes and characterized by unique organizational principles. Moreover, each family encompasses a group of objects typically found in its respective setting. This section outlines the intended function of organizing tabletop environments within each task family and enumerates the associated objects included in our datasets. We also provide a task family overview in Table \ref{tab:task_diversity_overview} to demonstrate the diversity of the tasks.

\noindent
\subsubsection{Study Desk}
\begin{itemize}
    \item \textbf{Purpose}: To create a functional study space. This involves strategically placing essential items for studying within easy reach. For example, input devices (e.g., keyboard or notebook) are placed near the table's edge closest to the user, while output devices (e.g., monitor) are positioned at the rear to protect eyesight. Objects not directly related to study activities are also thoughtfully arranged; for instance, a mug is placed on the right side of the table for a right-handed user for easy access.
    \item \textbf{Objects}: monitor, keyboard, laptop, notepad, book, mouse, pen, lamp, mug, tissue box, glasses, toy.
    \item \textbf{Training \& Testing datasets}: Table \ref{tab:study_table_training} contains the training examples, each consisting of user instructions, object lists, and FORM layouts. The testing dataset, which includes user instructions and object lists, is presented in Table \ref{tab:study_table_testing}.
\end{itemize}

\begin{longtable}{|p{2cm}|p{3.5cm}|p{5cm}|p{3.5cm}|p{2cm}|}
\caption{Task Dataset Overview} \label{tab:task_diversity_overview} \\
\hline
\textbf{Task Family} & \textbf{Common Objects} & \textbf{Functions} & \textbf{Personal Preferences} & \textbf{Object Range} \\
\hline
\endfirsthead
\multicolumn{5}{c}%
{\tablename\ \thetable\ -- \textit{Continued from previous page}} \\
\hline
\textbf{Task Family} & \textbf{Common Objects} & \textbf{Functions} & \textbf{Personal Preferences} & \textbf{Object Range} \\
\hline
\endhead
\hline \multicolumn{5}{r}{\textit{Continued on next page}} \\
\endfoot
\hline
\endlastfoot

Study Desk & Monitor, toy, mug, lamp, glasses, tissue box, notepad, pen, mouse, book, keyboard, laptop & \begin{itemize}
    \item Study with laptop
    \item Read books
    \item Use external monitor, keyboard, mouse, and other electronic devices
    \item Work with two notepads
    \item Work with multiple screens simultaneously
\end{itemize}
& \begin{itemize}
    \item Left-handed
    \item Easy access to books
    \item Prefer notepad over laptop
    \item Prefer laptop over other electronic devices
\end{itemize} & 4-10 \\
\hline
Dining Table & Rice bowl, spoon, fork, chopsticks, small plate, serving plate, napkin, baby bowl, baby plate, ramen bowl, seasoning, glass, medium plate, baby spoon, baby cup, knife & \begin{itemize}
    \item Set up Western dining table
    \item Set up Chinese dining table
    \item Set up ramen dining table
    \item Single diner setup
    \item Two diner setup 
    \item Four diner setup
\end{itemize}
& 
\begin{itemize}
    \item Share main dishes
    \item Dine with kids
    \item Sit on the same side
    \item Left-handed
\end{itemize}
& 9 - 24 \\
\hline
Coffee Table & Coffee pot, keys, tray, laptop, ashtray, iPad, tea cup, remote controller, beverage, snack bowl, vase, tea pot, candle, glasses, chess board, coffee cup, notepad & 
\begin{itemize}
    \item Tidy up the coffee table
    \item Party setup with snacks
    \item Chinese chess setup
    \item Temporary study space
    \item Afternoon tea setup
    \item Snack sharing setup
\end{itemize}& 
\begin{itemize}
    \item Create more space
    \item Easy access to certain objects
    \item Romantic setting
    \item Shared objects
    \item Maximize storage
\end{itemize}
& 7 - 13 \\
\end{longtable}

\begin{table}[htbp]
\centering
\caption{Training examples for arranging study desks.}
\label{tab:study_table_training}
\begin{tabular}{c|p{3cm}|p{3cm}|c}
\hline
\textbf{Index} & \textbf{User Instruction} & \textbf{Object List} & \textbf{FORM Arrangement} \\ 
\hline
1 & Could you please set up a study desk for me? I will mainly be using it to work on my laptop. I probably need the books for quick reference. & laptop, book\_1, book\_2, book\_3, book\_4, lamp, mouse & \includegraphics[width=6cm, height=1.8cm, valign=t]{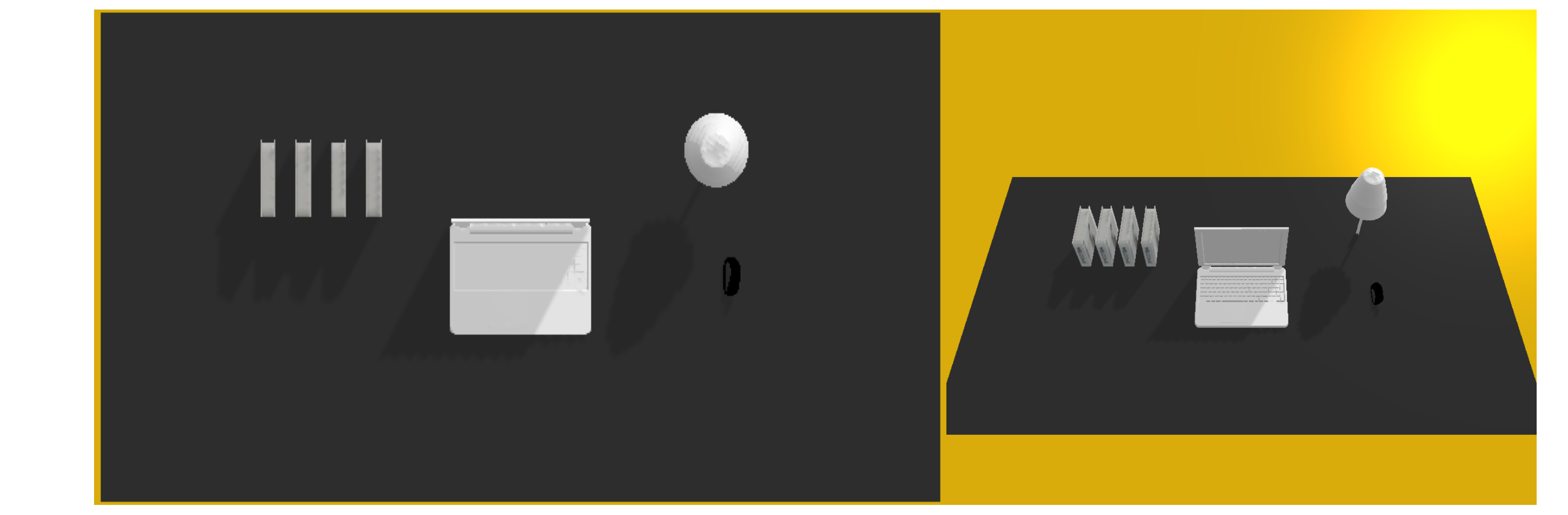} \\ 
\hline
2 & Could you please set up a study desk for me? I need to work on a laptop connecting to an external monitor. & monitor, laptop, lamp, mouse, mug & \includegraphics[width=6cm, height=1.8cm, valign=t]{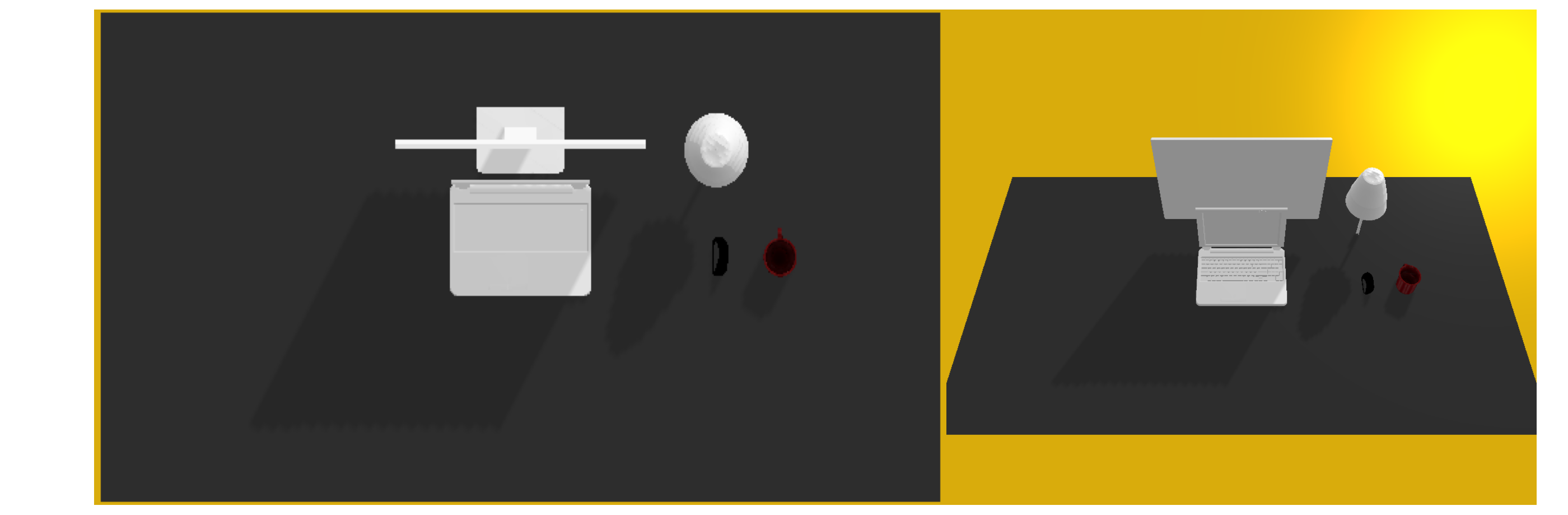} \\ 
\hline
3 & Could you please set up a study desk for me? I need to work on a laptop connecting to an external keyboard. & laptop, keyboard, lamp, mouse, mug & \includegraphics[width=6cm, height=1.8cm, valign=t]{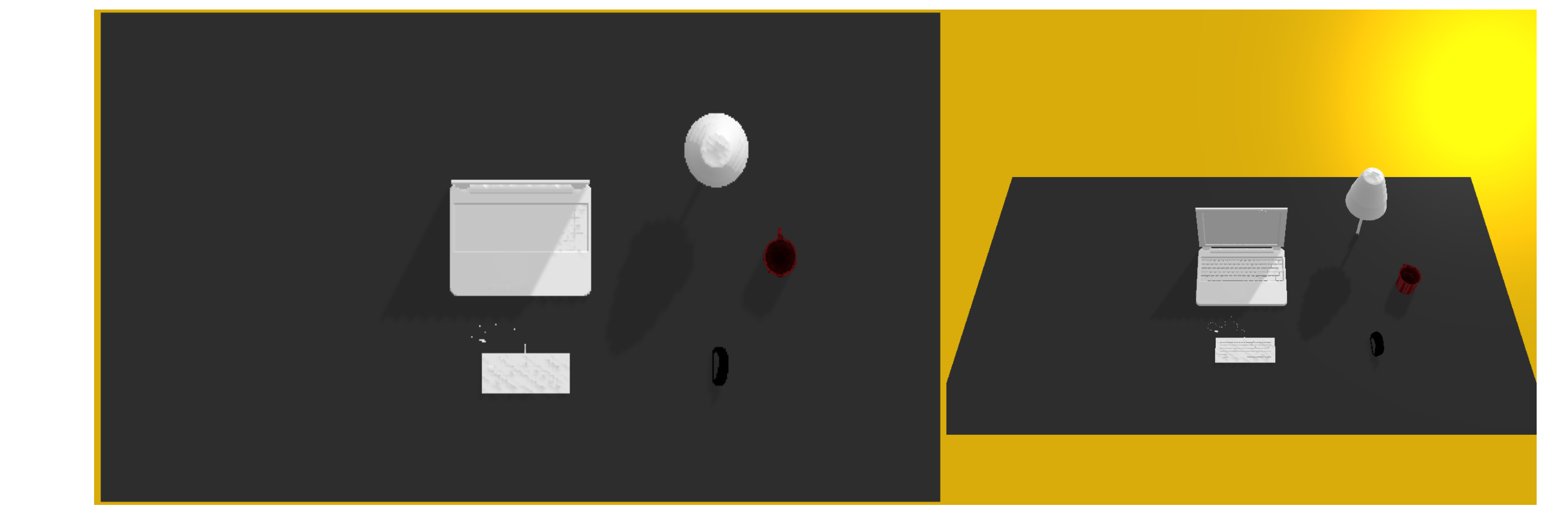} \\ 
\hline
4 & Could you please set up a study desk for me? I need to work on two laptops at the same time. & laptop\_1, laptop\_2, book\_1, book\_2, book\_3, book\_4, lamp & \includegraphics[width=6cm, height=1.8cm, valign=t]{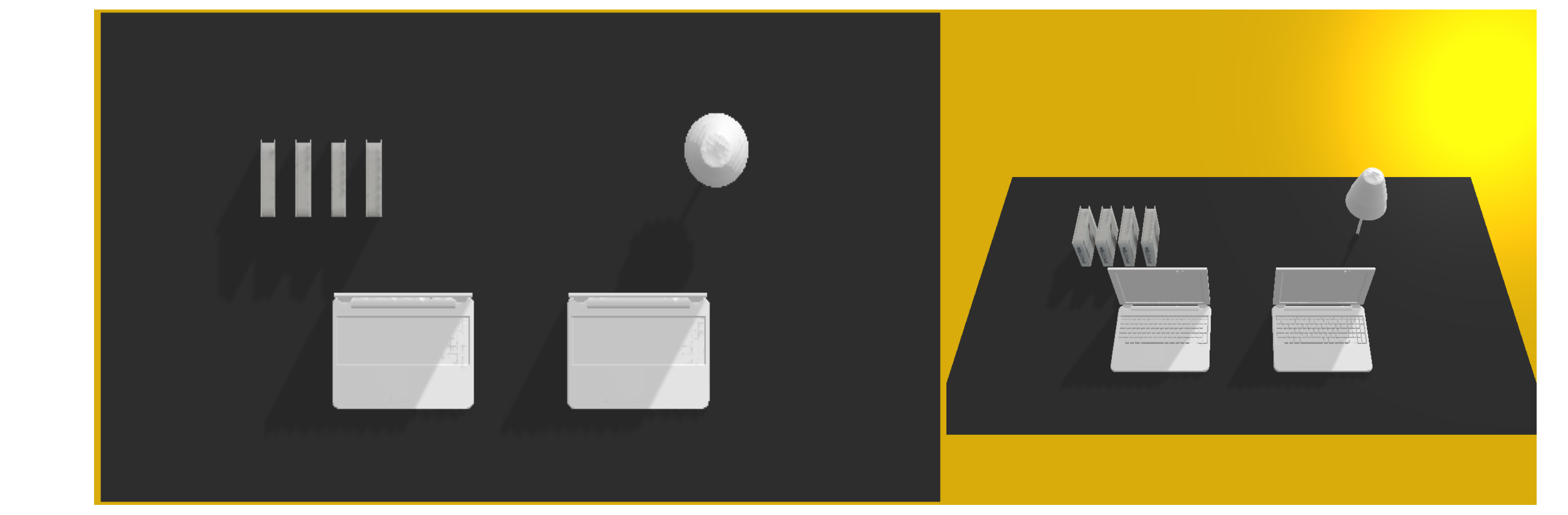} \\ 
\hline
5 & Could you please set up a study desk for me? I need to work on my notepad. & notepad, pen, book\_1, book\_2, book\_3, book\_4, lamp & \includegraphics[width=6cm, height=1.8cm, valign=t]{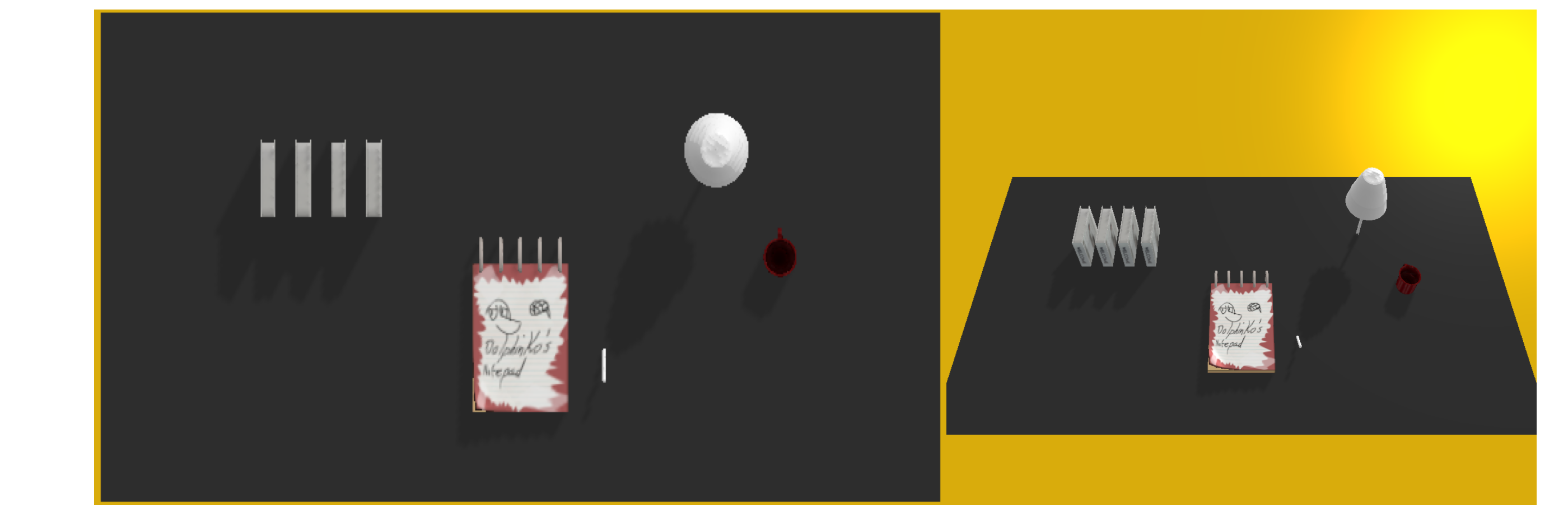} \\ 
\hline
\end{tabular}
\end{table}

\begin{table}[htbp]
\centering
\caption{Test cases for arranging study desks.}
\label{tab:study_table_testing}
\begin{tabular}{c|p{7cm}|p{7cm}}
\hline
\textbf{Index} & \textbf{User Instruction} & \textbf{Object List}  \\ 
\hline
1 & Could you please set up a study desk for me? I will mainly be using it to work on my laptop. I probably need the books for quick reference. & laptop, book\_1, book\_2, book\_3, book\_4, mug \\
\hline
2 & Would you be able to set up a study desk for me that's suitable for working on my computer? & monitor, keyboard, mouse, lamp, mug \\
\hline
3 & Could you set up a study desk for me that accommodates my laptop with an external monitor and keyboard connected? & monitor, laptop, keyboard, mouse \\
\hline
4 & Could you please arrange a study desk for me? I predominantly use my notepad for work. & notepad, pen, tissue\_box, mug, lamp \\
\hline
5 & Could you please set up a study desk for me with enough space to work on two notepads at the same time? I probably need the books for quick reference. & book\_1, book\_2, book\_3, book\_4, lamp, notepad\_1, notepad\_2, pen \\
\hline
6 & Could you please set up a study desk for me? I will mainly be using it to work on my laptop. & laptop, mouse, lamp, mug, tissue\_box \\
\hline
7 & Could you please arrange a study desk for me? I predominantly use my notepad for work. & monitor, notepad, pen, mug, lamp \\
\hline
8 & Could you please set up a study desk for me with enough space to work on a laptop and a notepad at the same time? & monitor, lamp, laptop, notepad, pen, tissue\_box, toy, mug \\
\hline
9 & Could you set up a study desk for me that accommodates my laptop with an external monitor and keyboard connected? & monitor, laptop, keyboard, mouse, glasses, mug, tissue\_box \\
\hline
10 & Could you please set up a study desk for me with enough space to work on a laptop and a notepad at the same time? I probably need the books for quick reference. & monitor, laptop, book\_1, book\_2, book\_3, book\_4, notepad, pen, mug, tissue\_box \\
\hline
\end{tabular}
\end{table}

\begin{table}[t!]
\centering
\caption{Training examples for arranging coffee tables.}
\label{tab:coffee_table_training}
\begin{tabular}{c|p{3cm}|p{3cm}|c}
\hline
\textbf{Index} & \textbf{User Instruction} & \textbf{Object List} & \textbf{FORM Arrangement} \\ 
\hline
1 & Could you please tidy up the coffee table? & tray, keys, remote\_controller, vase, candle\_1, candle\_2, magazine\_1, magazine\_2, magazine\_3, magazine\_4, & \includegraphics[width=6cm, height=1.8cm, valign=t]{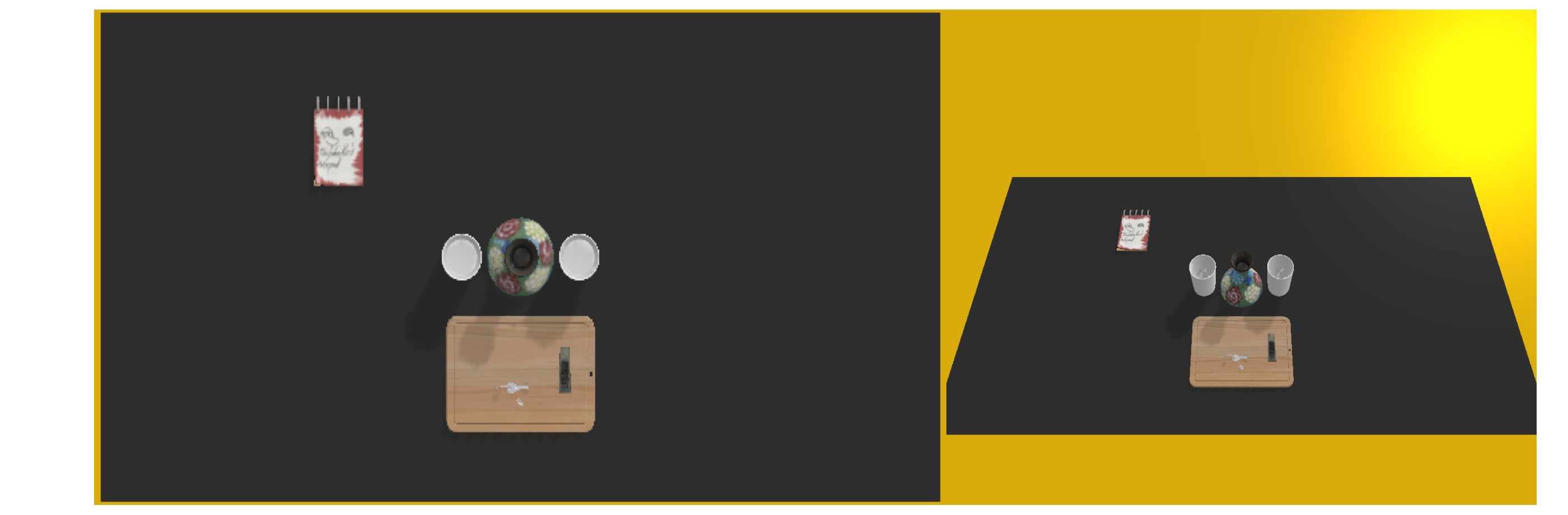} \\ 
\hline
2 & Could you please help me set up a coffee table for party? I have two bowls of snack to share. & Tray, keys, remote\_controller, snack\_bowl\_1, snack\_bowl\_2 & \includegraphics[width=6cm, height=1.8cm, valign=t]{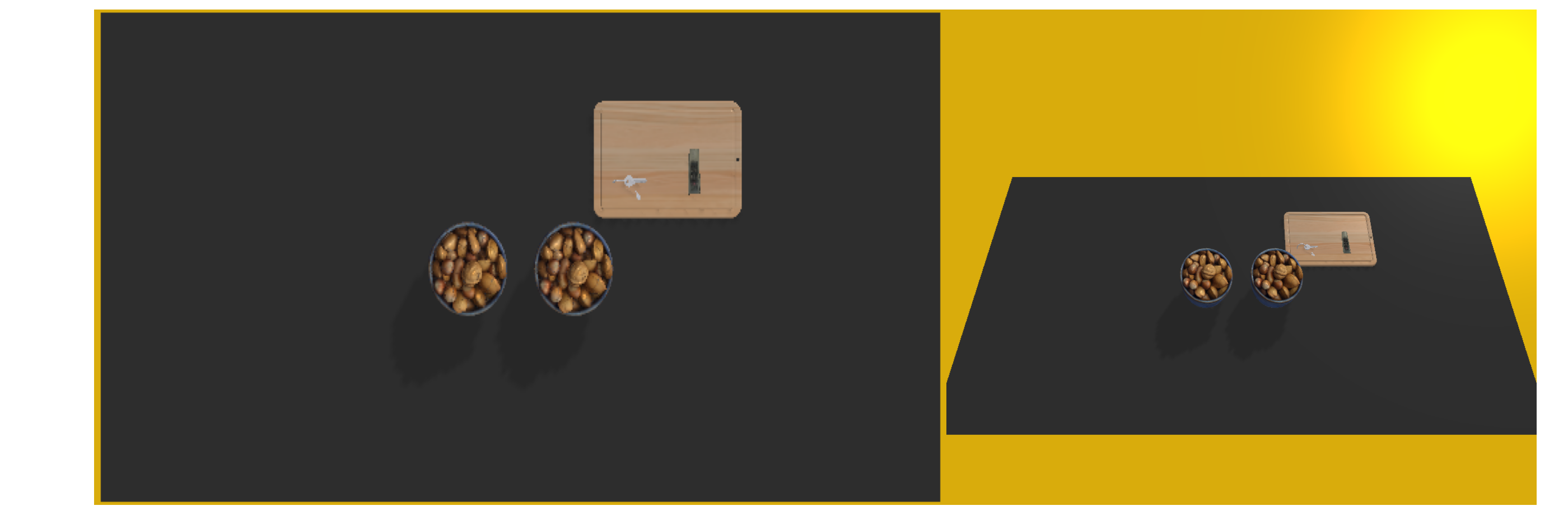} \\ 
\hline
3 & Could you assist me in arranging a coffee table for a tea and cake session for two? & coffee\_pot, coffee\_cup\_1, coffee\_cup\_2, cake\_plate\_1, cake\_plate\_2, magazine\_1, magazine\_2, magazine\_3, magazine\_4& \includegraphics[width=6cm, height=1.8cm, valign=t]{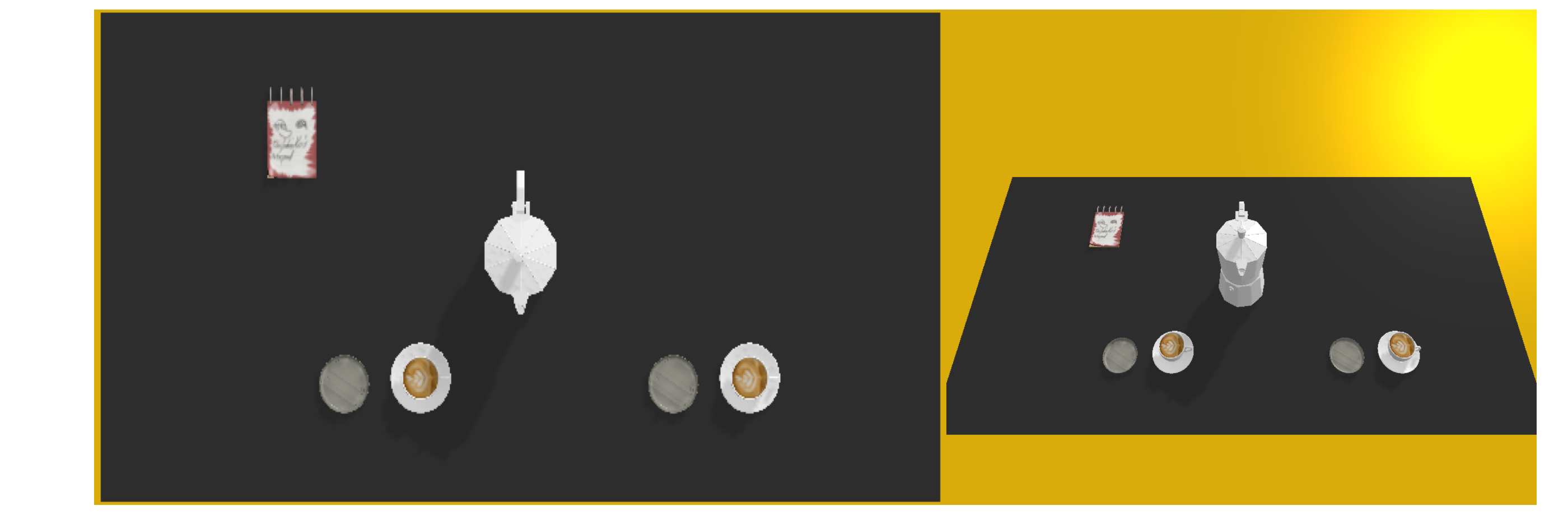} \\ 
\hline
4 & Could you please help me make some space on the coffee table? I want to read the magazine. & tray, keys, remote\_controller, vase, candle\_1, candle\_2, magazine & \includegraphics[width=6cm, height=1.8cm, valign=t]{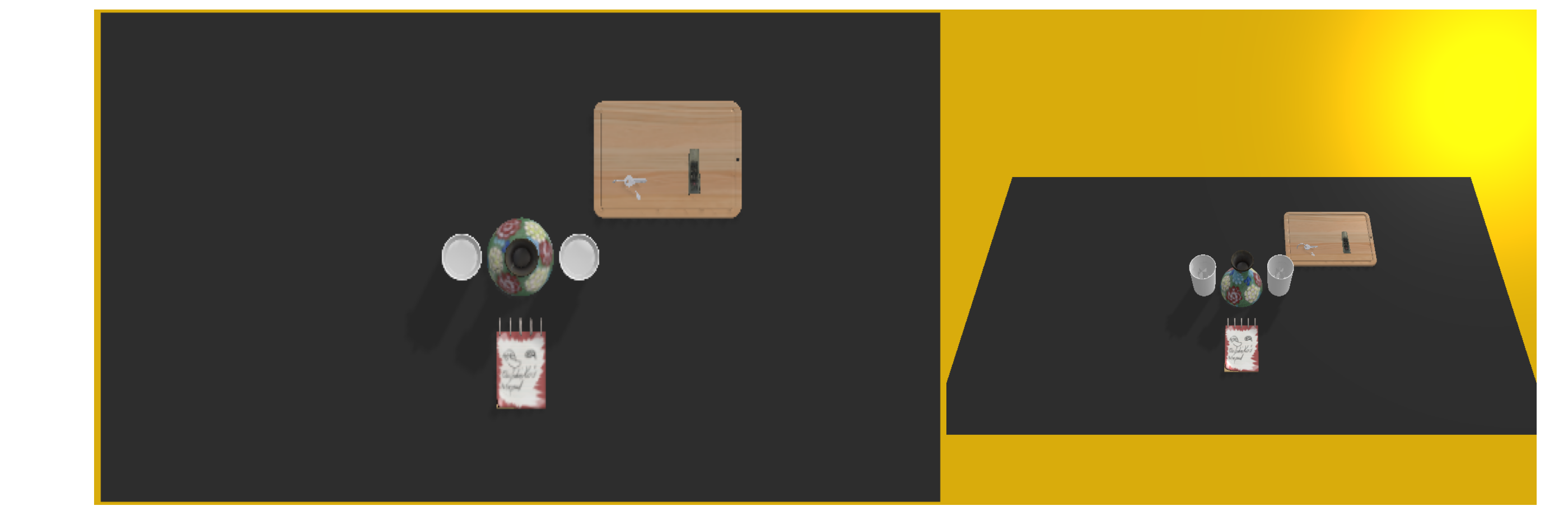} \\ 
\hline
5 & Could you please tidy up the coffee table to create more space, and make sure the tray is within easy reach so I can place my keys there? & Tray, keys, remote\_controller, vase\_1, vase\_2, vase\_3, magazine\_1, magazine\_2, magazine\_3, magazine\_4  & \includegraphics[width=6cm, height=1.8cm, valign=t]{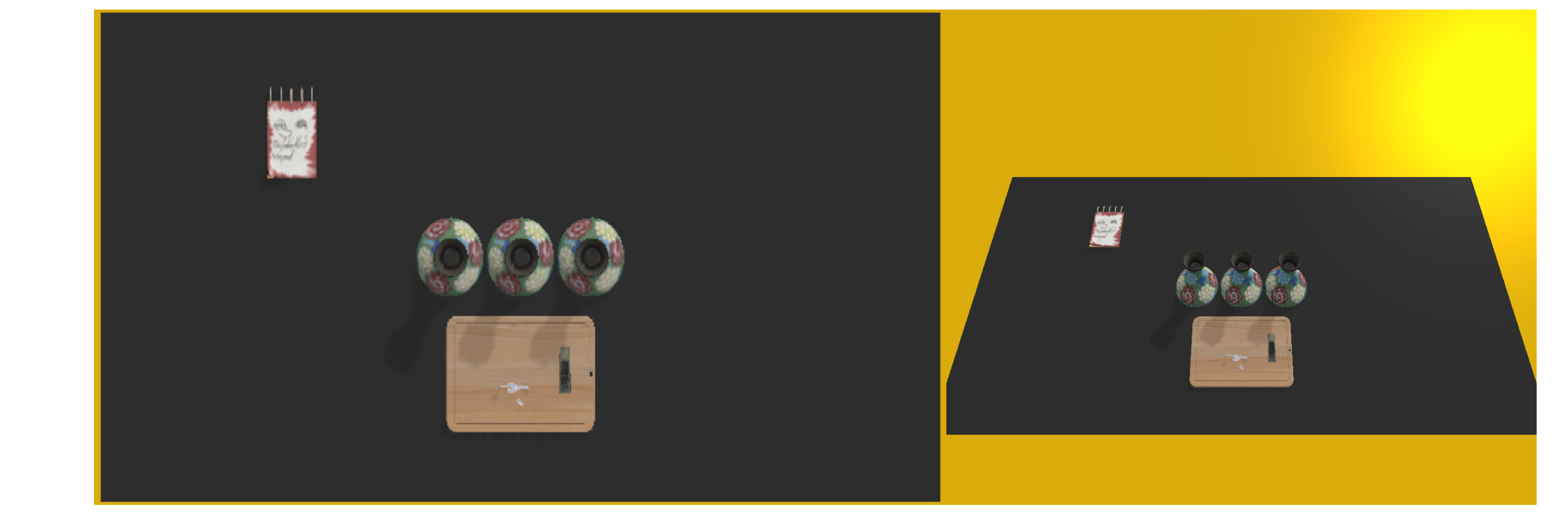} \\ 
\hline
\end{tabular}
\end{table}

\begin{table}[!h]
\centering
\caption{Test cases for arranging coffee tables.}
\label{tab:coffee_table_testing}
\begin{tabular}{c|p{7cm}|p{7cm}}
\hline
\textbf{Index} & \textbf{User Instruction} & \textbf{Object List}  \\ 
\hline
1 & Could you please tidy up the coffee table. & vase, candle\_1, candle\_2, tray, keys, glasses, remote\_controller \\
\hline
2 & Could you please tidy up the coffee table to create more space, and make sure the tray is within easy reach so I can place my keys there? & vase\_1, vase\_2, notepad\_1, notepad\_2, notepad\_3, notepad\_4, tray, keys, remote\_controller \\
\hline
3 & Please set up the coffee table for the party, arranging snacks and drinks so everyone can easily access them, and ensure other items take up minimal space. & ashtray, notepad\_1, notepad\_2, notepad\_3, notepad\_4, beverage\_1, beverage\_2, beverage\_3, beverage\_4, beverage\_5, beverage\_6, snack\_bowl\_1, snack\_bowl\_2 \\
\hline
4 & Could you please arrange the coffee table for a game of Chinese chess, along with coffee for two? & vase, candle\_1, candle\_2, chess\_board, coffee\_cup\_1, coffee\_cup\_2 \\
\hline
5 & Could you please clear the coffee table for me? I need to use it to work on my notenotepad. & vase, notepad\_1, notepad\_2, notepad\_3, notepad\_4, notenotepad, coffee\_cup \\
\hline
6 & Could you set up the coffee table for tea time for two people, please? & tea\_pot, tea\_cup\_1, tea\_cup\_2, tray, keys, glasses, remote\_controller \\
\hline
7 & Could you please set up the coffee table for a romantic setting with candles? & vase, candle\_1, candle\_2, candle\_3, candle\_4, notepad\_1, notepad\_2, notepad\_3, notepad\_4 \\
\hline
8 & Could you please set the coffee table for two people to enjoy coffee? & coffee\_pot, coffee\_cup\_1, coffee\_cup\_2, notepad\_1, notepad\_2, notepad\_3, notepad\_4, tray, keys, glasses, remote\_controller \\
\hline
9 & Please set up the coffee table with snacks for sharing and coffee for two. & snack\_bowl\_1, snack\_bowl\_2, snack\_bowl\_3, snack\_bowl\_4, tea\_cup\_1, tea\_cup\_2, tray, keys, glasses, remote\_controller \\
\hline
10 & Could you please clear the coffee table? I need to use it for working on my laptop. & vase, laptop, notepad\_1, notepad\_2, notepad\_3, notepad\_4, tray, keys, glasses, remote\_controller \\
\hline
\end{tabular}
\end{table}

\noindent
\subsubsection{Coffee Table}

\begin{table}[bp]
\centering
\caption{Training examples for arranging dining tables.}
\label{tab:dining_table_training}
\begin{tabular}{c|p{4.5cm}|p{4.5cm}|c}
\hline
\textbf{Index} & \textbf{User Instruction} & \textbf{Object List} & \textbf{FORM Arrangement} \\ 
\hline
1 & Could you please arrange a dining table for two people? & serving\_plate\_1, napkin\_1, fork\_1, knife\_1, spoon\_1, serving\_plate\_2, napkin\_2, fork\_2, knife\_2, spoon\_2 & \includegraphics[width=7cm, height=2cm, valign=t]{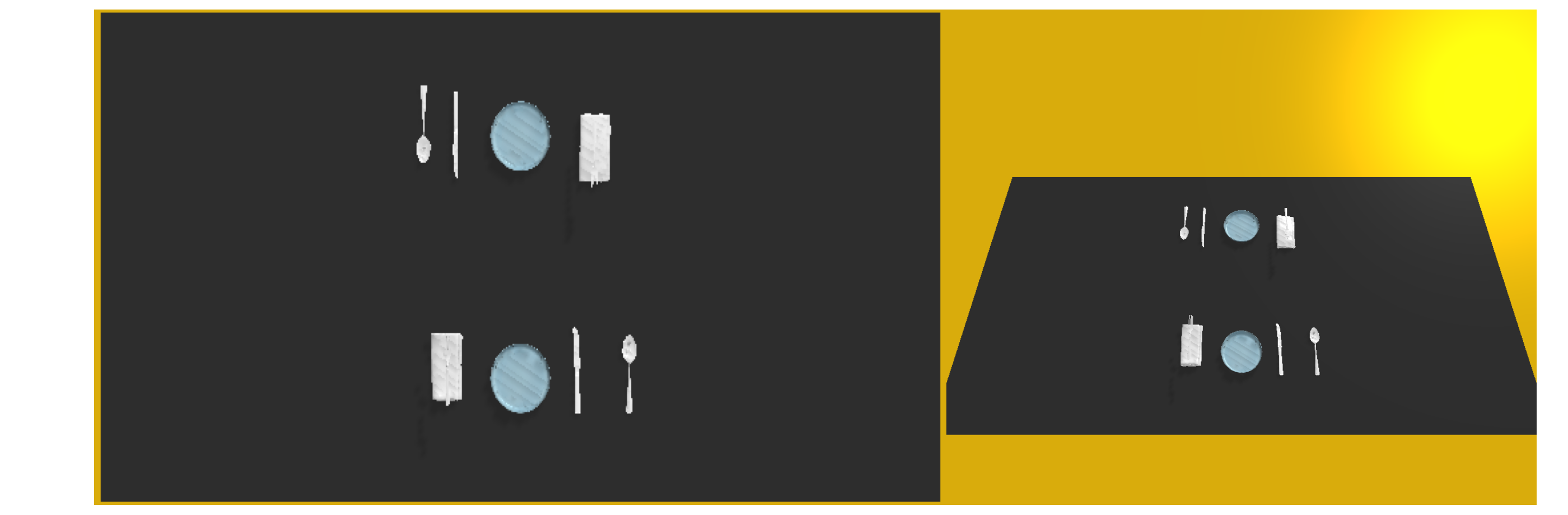} \\ 
\hline
2 & Please prepare a Chinese-style dining table for two guests. & medium\_plate\_1, medium\_plate\_2, small\_plate\_1, small\_plate\_2, rice\_bowl\_1, rice\_bowl\_2, chopsticks\_1, chopsticks\_2, spoon\_1, spoon\_2 & \includegraphics[width=7cm, height=2cm, valign=t]{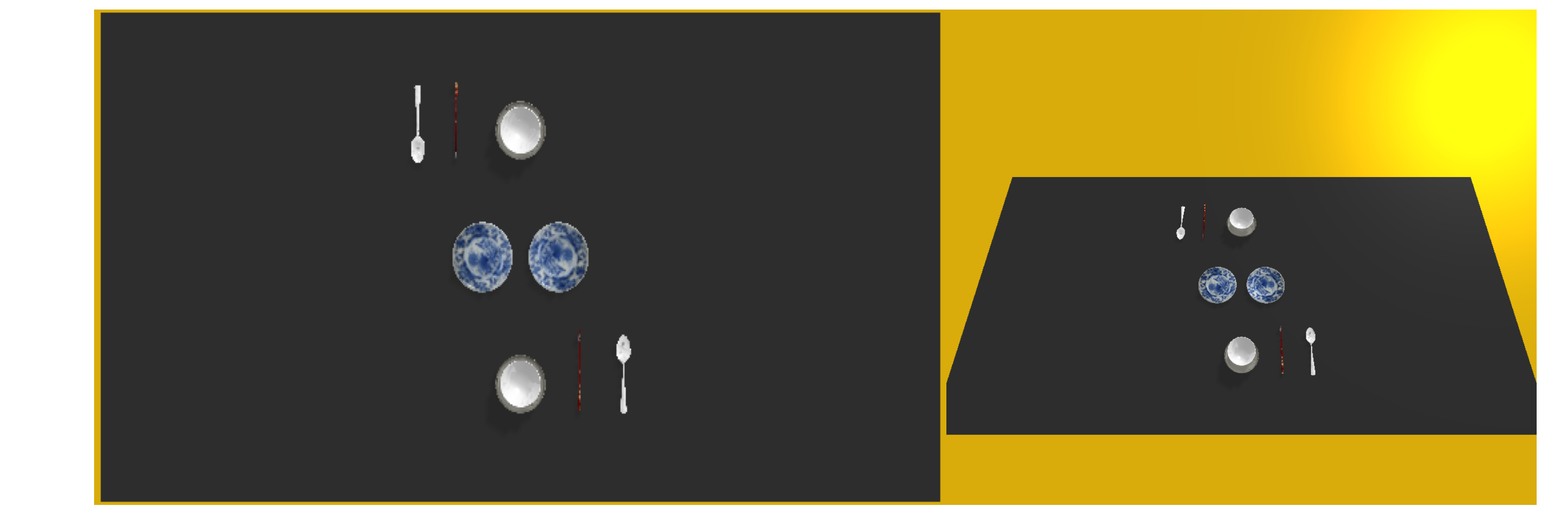} \\ 
\hline
3 & Could you please arrange a dining table for two? We would like to sit side by side? & serving\_plate\_1, napkin\_1, fork\_1, knife\_1, spoon\_1, serving\_plate\_2, napkin\_2, fork\_2, knife\_2, spoon\_2 & \includegraphics[width=7cm, height=2cm, valign=t]{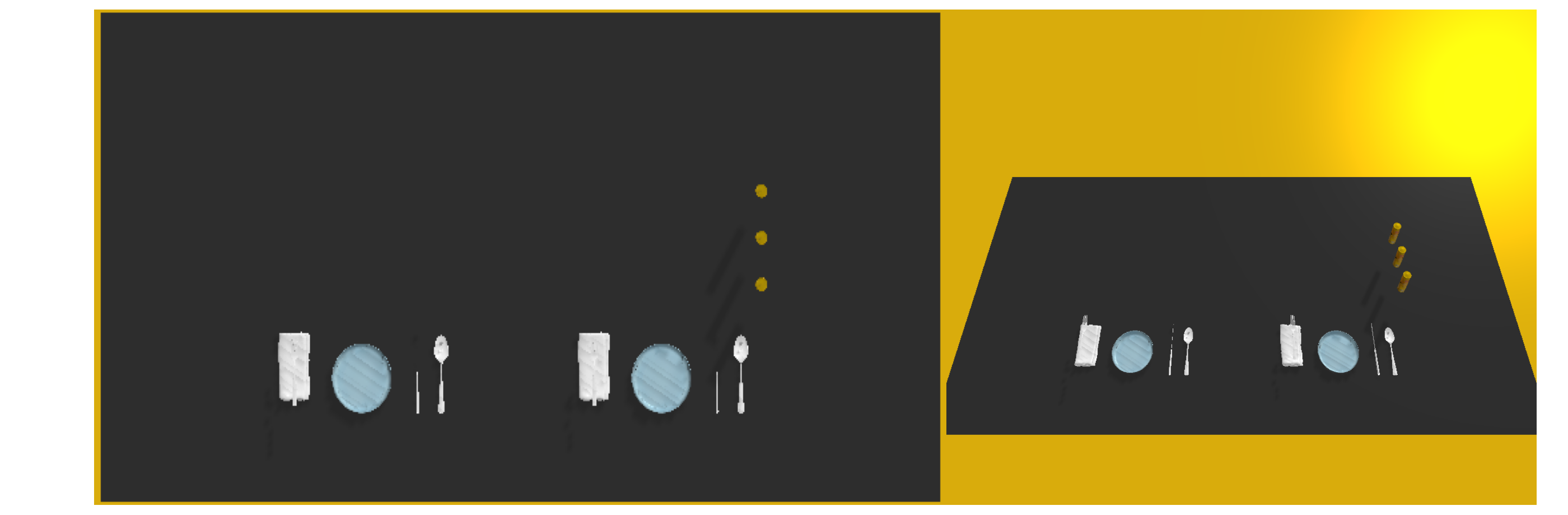} \\ 
\hline
4 & Could you please set up the ramen dining table for two, ensuring that the seating accommodates one left-handed diner? & ramen\_bowl\_1, chopsticks\_1, spoon\_1, glass\_1, ramen\_bowl\_2, chopsticks\_2, spoon\_2, glass\_2, seasoning\_1, seasoning\_2, seasoning\_3 & \includegraphics[width=7cm, height=2cm, valign=t]{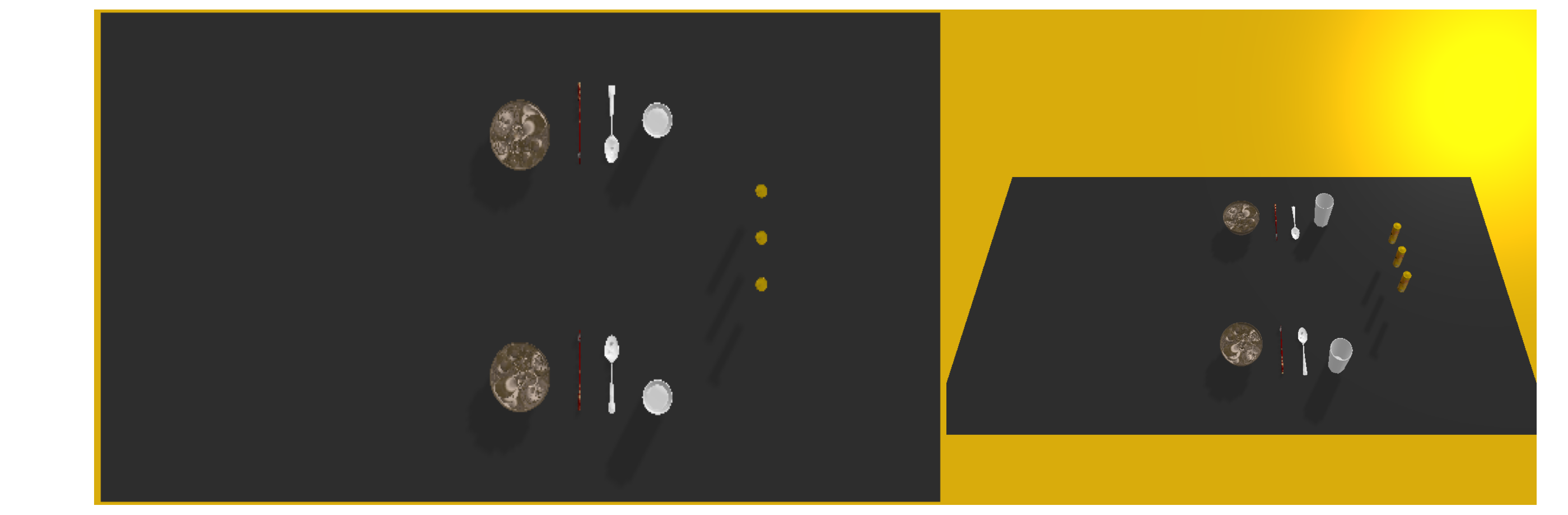} \\ 
\hline
5 & Instruction: Could you please set up a dining table for two, with the setup for sharing the main dishes? & medium\_plate\_1, medium\_plate\_2, serving\_plate\_1, napkin\_1, fork\_1, knife\_1, spoon\_1, serving\_plate\_2, napkin\_2, fork\_2, knife\_2, spoon\_2, seasoning\_1, seasoning\_2, seasoning\_3 & \includegraphics[width=7cm, height=2cm, valign=t]{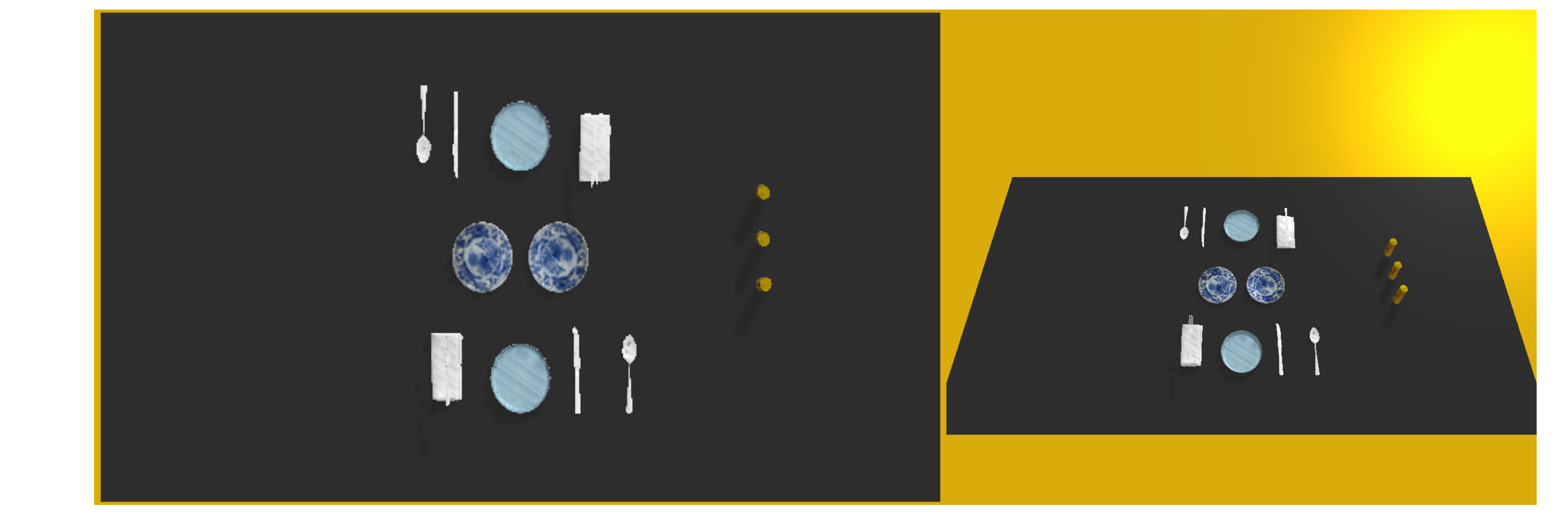} \\ 
\hline
\end{tabular}
\end{table}

\begin{table}[hp]
\centering
\caption{Test cases for arranging dining tables.}
\label{tab:dining_table_testing}
\begin{tabular}{c|p{7cm}|p{7cm}}
\hline
\textbf{Index} & \textbf{User Instruction} & \textbf{Object List}  \\ 
\hline
1 & Could you please arrange a dining table for two people? & serving\_plate\_1, napkin\_1, fork\_1, knife\_1, spoon\_1, glass\_1, serving\_plate\_2, napkin\_2, fork\_2, knife\_2, spoon\_2, glass\_2 \\
\hline
2 & Please prepare a Chinese-style dining table for two guests. & medium\_plate\_1, medium\_plate\_2, medium\_plate\_3, medium\_plate\_4, small\_plate\_1, small\_plate\_2, rice\_bowl\_1, rice\_bowl\_2, chopsticks\_1, chopsticks\_2, spoon\_1, spoon\_2 \\
\hline
3 & Could you please arrange a dining table for two? We would like to sit side by side. & serving\_plate\_1, napkin\_1, fork\_1, knife\_1, spoon\_1, glass\_1, serving\_plate\_2, napkin\_2, fork\_2, knife\_2, spoon\_2, glass\_2 \\
\hline
4 & Please prepare a table for one, set up for dining on ramen. & ramen\_bowl, chopsticks, spoon, medium\_plate\_1, medium\_plate\_2, seasoning\_1, seasoning\_2, seasoning\_3, seasoning\_4 \\
\hline
5 & Could you please arrange a dining table for a parent and a child, with seating on the same side to make it easier for me to tend to my child? & serving\_plate, napkin, fork, knife, spoon, glass, baby\_plate, baby\_bowl, baby\_spoon, baby\_cup, seasoning\_1, seasoning\_2, seasoning\_3 \\
\hline
6 & Could you please set up a dining table for two, ensuring that the seating accommodates one left-handed diner? & serving\_plate\_1, napkin\_1, fork\_1, knife\_1, spoon\_1, glass\_1, serving\_plate\_2, napkin\_2, fork\_2, knife\_2, spoon\_2, glass\_2 \\
\hline
7 & Could you please arrange a ramen dining table for a parent and a child, with seating on the same side to make it easier for me to tend to my child? & baby\_bowl, baby\_spoon, baby\_cup, ramen\_bowl, chopsticks, spoon, medium\_plate\_1, medium\_plate\_2, glass, seasoning\_1, seasoning\_2, seasoning\_3, seasoning\_4 \\
\hline
8 & Please prepare a Chinese-style dining table for two, with the arrangement made for shared main dishes. & medium\_plate\_1, medium\_plate\_2, medium\_plate\_3, medium\_plate\_4, medium\_plate\_5, medium\_plate\_6, small\_plate\_1, small\_plate\_2, rice\_bowl\_1, rice\_bowl\_2, chopsticks\_1, chopsticks\_2, spoon\_1, spoon\_2 \\
\hline
9 & Could you please set up a ramen dining table for two, with the setup for sharing the main dishes? & medium\_plate\_1, medium\_plate\_2, medium\_plate\_3, medium\_plate\_4, ramen\_bowl\_1, chopsticks\_1, spoon\_1, glass\_1, ramen\_bowl\_2, chopsticks\_2, spoon\_2, glass\_2, seasoning\_1, seasoning\_2, seasoning\_3 \\
\hline
10 & Please arrange a dining table for four, with seating on two opposite sides of the table. & serving\_plate\_1, napkin\_1, fork\_1, knife\_1, spoon\_1, glass\_1, serving\_plate\_2, napkin\_2, fork\_2, knife\_2, spoon\_2, glass\_2, serving\_plate\_3, napkin\_3, fork\_3, knife\_3, spoon\_3, glass\_3, serving\_plate\_4, napkin\_4, fork\_4, knife\_4, spoon\_4, glass\_4 \\
\hline
\end{tabular}
\end{table}

\begin{itemize}
    \item \textbf{Purpose}: The coffee table is organized either to tidy away small items or to set up for specific events. If no specific activity is required, the goal is to arrange the space to easily store and access small items. When preparing for activities, object placement considers the number of participants and the nature of the activity, ensuring items like shared snacks are centrally located for easy accessibility.
    \item \textbf{Objects}: candle, vase, book, laptop, tray, keys, remote controller, glasses, ashtray, snack bowl, beverage, coffee pot, teapot, tea cup, coffee cup, cake plate, board game.
    \item \textbf{Training \& Testing datasets}: Table \ref{tab:coffee_table_training} contains the training examples, each consisting of user instructions, object lists, and FORM layouts. The testing dataset, which includes user instructions and object lists, is presented in Table \ref{tab:coffee_table_testing}.
\end{itemize}

\noindent
\subsubsection{Dining Table}

\begin{itemize}
    \item \textbf{Purpose}: To prepare the dining table based on the specific dining style and the number of diners, accommodating any special requests (e.g., preferences for left-handed individuals). Utensils and tablewares are set according to conventional dining etiquette for the chosen dining style.
    \item \textbf{Objects}: serving plate, fork, napkin, spoon, knife, glass, ramen bowl, medium plate, chopsticks, small plate, rice bowl, baby bowl, baby plate, baby cup, baby spoon, baby fork, seasoning bottle.
    \item \textbf{Training \& Testing datasets}: Table \ref{tab:dining_table_training} contains the training examples, each consisting of user instructions, object lists, and FORM layouts. The testing dataset, which includes user instructions and object lists, is presented in Table \ref{tab:dining_table_testing}.
\end{itemize}

%% file: appendix/07-robot_setup.tex
\subsection{Robot Setup}
\label{ssec:robot}

Our real-world setup, contains a Franka Emika Panda robot arm with a parallel gripper, mounted on a table, and an Intel Realsense D435 camera mounted on the table frame, pointing (approximately) 45 degrees down and giving a top-front view of the table. Our vision pipeline contains six steps: first, based on the calibrated camera intrinsics and extrinsics, we reconstruct a partial point cloud for the scene. Second, we crop the scene to exclude volumes outside the table (e.g., background drops, etc.) Third, we use a RANSAC-based algorithm to estimate the table plane, and extract all object point clouds on top of the table. Fourth, we use Mask R-CNN to detect all objects in the RGB space, and extract the corresponding point clouds. For objects that are not detected by the MaskRCNN, we first use DBScan to cluster their point clouds and then run the SegmentAnything model to extract their segmentations in 2D and, subsequently, the point clouds. Finally, we perform object completion by projecting and extruding the bottom surface for all detected objects down to the table.

Based on the reconstructed object models, we compute their corresponding 2D bounding boxes. For objects without a name (\ie, objects not detected by the Mask R-CNN but generated by the DBScan pipeline), we ask human users to name their categories. Next, we give all object names, shapes, and human instructions (\eg, set up a breakfast table for me) to our system and predict the target object poses.

Then, given the predicted target object poses, which contain 2D locations and orientations of each object, we search for an order for object placement so that there is no object collision during the re-arrangement based on their initial position. For simplicity, we only consider the order for directly moving objects to their target locations and do not consider strategies such as putting an object on the side to clear up any additional workspace. After fixing the object arrangement order, we use a hand-coded grasp sampler for top-down grasping poses by finding parallel surfaces using the reconstructed pointclouds, and use a scripted policy for reaching the object and placing the object, both from the top. The visualizations of segmentation results, generated orders, and grasping poses are included in the supplementary video.

%% file: appendix/08-adding_new_relationship.tex
\subsection{{Adding New Spatial Relationships}}

\model support \textbf{continual learning to integrate new spatial relations} and is adaptable to more complex arrangements. If an arrangement requires spatial relations not currently in our library, we can train a new diffusion model with synthetic data corresponding to the new relation. This approach allows \model to adapt to the complexities and nuances of real-world object arrangements.

\vspace{0.5em}
\textbf{Adding new abstract relation.} \label{EufD:adding_relation}
Users can simply integrate new abstract relations by the following two steps: 1) collecting training data, and 2) training the individual constraint diffusion solver. There are two ways to collect the training data, depending on the user case. 

\vspace{0.5em}
\begin{itemize}
    \item If the user can only \textbf{manually collect small amount} of training data (about 10), \model will learn a pose proposer that exactly replicates the specific arrangements. 
    \item If the user can \textbf{generate the script to create synthetic data} (about 1000), \model can learn a more robust pose proposer that generalizes to more number of objects.
\end{itemize}

\vspace{0.5em}
To demonstrate \model's capability to learn new abstract spatial relations, we add an experiment to include a new relation \textbf{sorted} based on width. We trained models with different numbers of training examples and evaluated their success rates in sorting objects of varying counts:

\input{fig-texts/sorted_relation}

\vspace{0.5em}
\textbf{Composing with existing relations}
The newly learned \textbf{sorted} relation can be combined with existing relations to create layouts that satisfy multiple spatial constraints. We visually demonstrate \model's compositionality through the following three test cases in Table \ref{tab:sorted_test_cases}, which compose the \textbf{sorted} with 2-4 existing relationships. The proposed layout is shown in Fig \ref{fig:composed_sorted}.

\input{fig-texts/sorted_test_cases}
\input{fig-texts/composed_sorted}

\vspace{0.5em}
In summary, \model's design addresses the concerns raised by embracing continual learning and adaptability. The system's ability to integrate new spatial relations and perform behavior cloning ensures that the subtleties of real-world object arrangements are not neglected.

%% file: fig-texts/sorted_relation.tex
\begin{table}[ht]\label{tab:sorted_relation}
\centering
\caption{Success Rate for Sorting N objects (out of 10 trials)}
\begin{tabular}{m{7cm}cccccc}
\toprule
\textbf{Training data} & \multicolumn{6}{c}{\textbf{Success Rate for Sorting N objects}} \\ \cmidrule{2-7} 
                       & \textbf{N = 3} & \textbf{N = 4} & \textbf{N = 5} & \textbf{N = 6} & \textbf{N = 7} & \textbf{N = 8} \\ \midrule
10 samples, N = 3 objects, manual collection & 8 & 2 & 1 & 0 & 0 & 0 \\ \midrule
100 samples, N = 3 objects, synthetic data & 10 & 9 & 3 & 2 & 0 & 0 \\ \midrule
1000 samples, N = 3, 4, 5 objects, synthetic data & 10 & 10 & 10 & 10 & 9 & 9 \\ \bottomrule
\end{tabular}
\end{table}

%% file: fig-texts/sorted_test_cases.tex
\begin{table}[ht]
    \centering
    \begin{tabular}{>{\raggedright\arraybackslash}p{4cm}|p{4cm}|p{7cm}}
        \hline
        \textbf{Cases} & \textbf{Objects List} & \textbf{Abstract Relationship} \\
        \hline
        3 plates sorted at the center of the table & medium\_plate\_1, medium\_plate\_2, medium\_plate\_3 &
        \begin{tabular}[t]{@{}l@{}}
            - sorted(medium\_plate\_1, medium\_plate\_2, medium\_plate\_3) \\
            - central\_column(medium\_plate\_1, medium\_plate\_2, medium\_plate\_3) \\
            - central\_row(medium\_plate\_1, medium\_plate\_2, medium\_plate\_3) \\
        \end{tabular} \\
        \hline
        5 vases sorted in the center column of the table near the back edge & vase\_1, vase\_2, vase\_3, vase\_4, vase\_5 &
        \begin{tabular}[t]{@{}l@{}}
            - sorted(vase\_1, vase\_2, vase\_3, vase\_4, vase\_5) \\
            - near\_back\_edge(vase\_1, vase\_2, vase\_3, vase\_4, vase\_5) \\
            - central\_column(vase\_1, vase\_2, vase\_3, vase\_4, vase\_5) \\
        \end{tabular} \\
        \hline
        4 forks and 4 spoons, the 4 forks sorted on the left half of the table near the front edge, while the 4 spoons sorted on the right half of the table near the front edge & fork\_1, fork\_2, fork\_3, fork\_4, spoon\_1, spoon\_2, spoon\_3, spoon\_4 &
        \begin{tabular}[t]{@{}l@{}}
            - sorted(fork\_1, fork\_2, fork\_3, fork\_4) \\
            - near\_front\_edge(fork\_1, fork\_2, fork\_3, fork\_4) \\
            - left\_half(fork\_1, fork\_2, fork\_3, fork\_4) \\
            - sorted(spoon\_1, spoon\_2, spoon\_3, spoon\_4) \\
            - near\_front\_edge(spoon\_1, spoon\_2, spoon\_3, spoon\_4) \\
            - right\_half(spoon\_1, spoon\_2, spoon\_3, spoon\_4) \\
        \end{tabular} \\
        \hline
    \end{tabular}
    \caption{Summary of cases and object placement with their abstract relationships.} \label{tab:sorted_test_cases}
\end{table}

%% file: fig-texts/composed_sorted.tex
\begin{figure}[hptb]
    \centering
    \includegraphics[width=\linewidth]{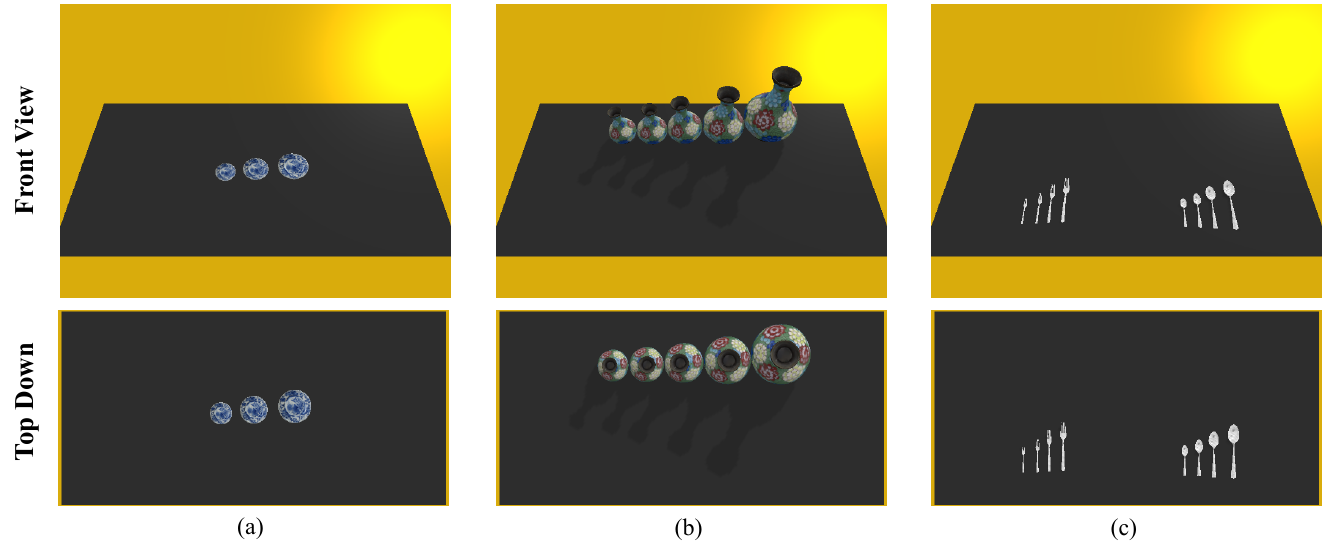}
    \caption{Composing the newly learned relationship \textbf{sorted} with the existing ones in the abstract relationship library.}
    \label{fig:composed_sorted}
\end{figure}

%% file: appendix/09-ablation_study.tex
\subsection{Ablation Study on Zero-shot \model}\label{appendix:zeroshot}

To study how does the user-provided examples influence the output, we conducted an ablation study to complement the human study results presented in the original manuscript. This additional study aimed to \textbf{quantify the impact of user-provided examples on the output} of the \model framework.

\vspace{0.5em}
\textbf{Ablation Study Methodology}: We removed the 5 training examples from \model and compared the resulting object arrangements to those generated with the inclusion of the examples. This comparison allowed us to isolate and assess the influence of user-defined preferences on the system's output.

\vspace{0.5em}
\textbf{Ablation Study Results}: The study revealed differences in the object arrangements in aspects of commonsense arrangements not explicitly specified in the sketch or instructions. For example, in the context of a study desk setup, the placement of a lamp varied significantly; without the examples, the lamp was placed next to a notepad or laptop, which is a logical but less personalized arrangement. With the examples, the lamp's placement was tailored to the user's specific preferences, such as proximity to the reading area or the user's dominant hand. The qualitative comparison is summarized in Table~\ref{tab:ablation_comparision}.

\vspace{0.5em}
The ablation study demonstrates that the 5 training examples play a crucial role in personalizing the commonsense arrangement knowledge embedded in the \model system. While the system can generate functional layouts without these examples, the inclusion of user-defined preferences leads to more personalized and user-centric arrangements. This finding underscores the importance of the two-stage preference integration process in capturing and reflecting individual user needs.

\input{fig-texts/ablation_comparison}

%% file: fig-texts/ablation_comparison.tex
\begin{table}[htbp]
\centering
\caption{Comparison Between SetItUp and its Zero-shot Variant. We present the difference in the generated layouts between the two methods.}
\label{tab:ablation_comparision}
\begin{tabular}{c p{7.5cm} p{7.5cm}}
\toprule
\textbf{Task Family} & \textbf{With 5 Training Examples} & \textbf{Without Training Examples}  \\ 
\midrule
Study Desk & Lamp placed based on reading area or user's dominant hand. Laptop positioned as an input/output device in the center. & Lamp placed next to notepad or laptop. Laptop classified as an output device and placed at the back. \\
\midrule
Coffee Table & Tray used for storage, placed at the back. Regular grid arrangement for object clusters.  & Tray used for serving coffee/tea, placed more centrally. Symmetrical alignments and arrangements for object clusters.\\
\midrule
Dining Table & Fork on the napkin, bowl on the plate. Objects spaced evenly on the same side (left\_half or right\_half).  & Fork next to the napkin, bowl next to the plate. Objects spaced evenly across the entire table. \\
\bottomrule
\end{tabular}
\end{table}

%% file: appendix/10-self-reflection.tex
\subsection{Quantitative Study on Self-Reflection}
\label{appendix:self_reflection}

We quantitatively study the effectiveness of the self-reflection mechanism. We report the number of iterations required to resolve inconsistencies for each test case. 

\vspace{0.5em}
For the three task families, inconsistency arises on a few instances for study desk and very rarely for dining table. In our analysis, we found that inconsistencies mostly arise from two primary causes:

\vspace{0.5em}
\begin{itemize}
    \item Spatial relations that are individually correct but jointly inconsistent.
    \item Ambiguous object categorization or objects not included in the initial sketch.
\end{itemize}

\vspace{0.5em}
Inconsistencies were resolved within three self-reflection iterations across all test cases. All the test cases that require the inconsistency resolution are listed in Table \ref{tab:self_reflection_stats}.

\input{fig-texts/self_reflection_stats}

%% file: fig-texts/self_reflection_stats.tex
\begin{table}[ht]
\centering
\caption{Test cases that require self-reflection.}
\label{tab:self_reflection_stats}
\begin{tabular}{m{2cm}m{2cm}m{6cm}m{5cm}}
\toprule
\textbf{Test Cases} & \textbf{\# of Self-Reflections} & \textbf{Inconsistency} & \textbf{Causes} \\
\midrule
Study desk, test case \#2 & 3 & ("horizontally aligned", "monitor", "laptop"), yet ("near the back edge", "monitor") and ("table\_center", "laptop") & Ambiguity in the functional role of the laptop: laptop can either be an output or I/O device. If output device, it aligns with monitor horizontally, if I/O, it locates at the table center. \\
\midrule
Study desk, test case \#4 & 1 & ("near\_back\_edge", "lamp"), ("left\_of", "lamp", "laptop"), ("centered\_table", "laptop") & The arrangement for the category of laptop not explicit listed in the sketch. Unclear if the lamp is meant to illuminate the laptop or the entire table. \\
\midrule
Study desk, test case \#8 & 1 & ("near\_back\_edge", "laptop"), ("front\_half", "notepad"), ("horizontally\_aligned", "laptop", "notepad") & Ambiguity in the functional role of the laptop (output/I/O device). If output device, it aligns with monitor horizontally, if I/O, it aligns with the notepad. \\
\midrule
Coffee table, test case \#9 & 2 & ("near\_back\_edge", "notepad"), ("left\_of", "notepad", "draft\_paper"), ("near\_the\_front\_edge", "draft\_paper") & Ambiguity in the functional role of the notepad. Unclear if the notepad is actively used in the current activity. \\
\bottomrule
\end{tabular}
\end{table}

%% file: appendix/11-additional_stats_test.tex
\subsection{Additional Statistical Testing for Evaluation}

We conduct additional statistical testing on the result reported in Table \ref{tab:combined_comparison}, so as to validate if they are statistically significant. We perform the following statistical test on each task family:

\vspace{0.5em}
\begin{enumerate}
    \item Initial Testing: We performed a Repeated Measures ANOVA, which confirmed significant differences exist among the four methods compared in the study. The result is reported in Table \ref{tab:ANOVA}.
    \item Assumption Validation: Prior to further tests, we verified assumptions for using parametric testing, ensuring the data met the necessary criteria for subsequent analyses. The result is reported in Table \ref{tab:testOfSphericity}.
    \item Post Hoc Comparison: To examine pair-wise differences between \model and each baseline method, we utilized paired-sample t-tests. These tests demonstrated that the improvements by \model over the baseline methods are statistically significant. The result is reported in Table \ref{tab:postHocComparisons}.
\end{enumerate}

\begin{table}[hp]
\centering
\caption{Initial Testing: Repeat Measure ANOVA test on various task families.}
\label{tab:ANOVA}
{
\begin{tabular}{lrrrrr}
\toprule
Task Family & Sum of Squares & df & Mean Square & F & p  \\
			\cmidrule[0.4pt]{1-6}
Study Desk & $79.568$ & $3$ & $26.523$ & $241.908$ & $<$ .001***  \\
Coffee Table & $72.980$ & $3$ & $24.327$ & $163.834$ & $<$ .001***  \\
Dining Table & $141.455$ & $3$ & $47.152$ & $415.892$ & $<$ .001***  \\
			\bottomrule
		\end{tabular}
	}
\end{table}

\begin{table}[htpb]
\centering
\caption{Assumption Validation: Test of Sphericity on each task faimilies.}
\label{tab:testOfSphericity}
{
\begin{tabular}{lrrrrrrr}
\toprule
& Mauchly's W & Approx. X$^{2}$ & df & p-value & Greenhouse-Geisser $\epsilon$ & Huynh-Feldt $\epsilon$ & Lower Bound $\epsilon$ \\
\cmidrule[0.4pt]{1-8}
Study Desk & $0.725$ & $6.333$ & $5$ & $0.276$ & $0.833$ & $0.954$ & $0.333$  \\
Coffee Table & $0.655$ & $8.335$ & $5$ & $0.139$ & $0.767$ & $0.867$ & $0.333$  \\
Dining Table & $0.570$ & $11.077$ & $5$ & $0.110$ & $0.753$ & $0.849$ & $0.333$  \\
\bottomrule
\end{tabular}}
\end{table}

\begin{table}[htpb]
\centering
\caption{Post Hoc Comparisons: paired-sample t-test for all task families.}
\label{tab:postHocComparisons}
{
\begin{tabular}{clrrrrrr}  
\toprule
& \multicolumn{1}{c}{} & \multicolumn{5}{c}{95\% CI for Mean Difference}\\
\cline{3-8}
& &  & Mean Difference & Lower & Upper  & t & p$_{\text{holm}}$ \\
\cmidrule[0.4pt]{2-8}
\multirow{6}{*}{Study Desk} & Ours & Direct LLM Prediction & $1.477$ & $1.205$ & $1.749$  & $14.797$ & $<$ .001 \\
&  & End-to-end Diffusion Model & $2.600$ & $2.328$ & $2.872$ & $26.043$ & $<$ .001 \\
& & LLM + Diffusion & $0.823$ & $0.551$ & $1.095$ & $8.241$ & $<$ .001 \\
& Direct LLM Prediction & End-to-end Diffusion Model & $1.123$ & $0.851$ & $1.395$ & $11.246$ & $<$ .001 \\
& & LLM + Diffusion & $-0.655$ & $-0.927$ & $-0.383$ & $-6.556$ & $<$ .001 \\
& End-to-end Diffusion Model & LLM + Diffusion & $-1.777$ & $-2.049$ & $-1.505$ & $-17.802$ & $<$ .001 \\
\cmidrule[0.4pt]{2-8}
\multirow{6}{*}{Coffee Table} & Ours & Direct LLM Prediction & $1.627$ & $1.311$ & $1.944$ & $14.006$ & $<$ .001 \\
&  & End-to-end Diffusion Model & $2.541$ & $2.224$ & $2.857$ & $21.870$ & $<$ .001 \\
& & LLM + Diffusion & $1.314$ & $0.997$ & $1.630$ &  $11.307$ & $<$ .001 \\
& Direct LLM Prediction & End-to-end Diffusion Model & $0.914$ & $0.597$ & $1.230$ & $7.864$ & $<$ .001 \\
& & LLM + Diffusion & $-0.314$ & $-0.630$ & $0.003$  & $-2.699$ & $0.009$ \\
& End-to-end Diffusion Model & LLM + Diffusion & $-1.227$ & $-1.544$ & $-0.911$ & $-10.563$ & $<$ .001 \\
\cmidrule[0.4pt]{2-8}
\multirow{6}{*}{Dining Table} & Ours & Direct LLM Prediction & $2.459$ & $2.183$ & $2.736$ & $24.222$ & $<$ .001 \\
& & End-to-end Diffusion Model & $3.209$ & $2.933$ & $3.486$ & $31.610$ & $<$ .001 \\
& & LLM + Diffusion Model & $2.918$ & $2.642$ & $3.195$ & $28.744$ & $<$ .001 \\
& Direct LLM Prediction & End-to-end Diffusion Model & $0.750$ & $0.473$ & $1.027$ & $7.388$ & $<$ .001 \\
& $ $ & LLM + Diffusion Model & $0.459$ & $0.183$ & $0.736$ & $4.522$ & $<$ .001 \\
& End-to-end Diffusion Model & LLM + Diffusion Model & $-0.291$ & $-0.567$ & $-0.014$ & $-2.865$ & $0.006$ \\
\bottomrule
\end{tabular}
}
\end{table}

We first look into the repeated measure ANOVA in Table \ref{tab:ANOVA}. For \textit{Study Desk}, the F-value is $241.908$ with a very significant p-value ($< .001$), indicating strong statistical differences among the methods. Similarly, \textit{Coffee Table} and \textit{Dining Table} task families show extremely significant p-values with F-values of $163.834$ and $415.892$ respectively. There are significant differences among the methods across all task families, justifying further pairwise comparisons.

\vspace{0.5em}
Next, perform the test of Sphericity (Table \ref{tab:testOfSphericity}) to check if the variances of the differences between all possible pairs of groups are equal, which is necessary for the validity of Repeated Measures ANOVA. For \textit{Study Desk}, its Mauchly's W is $0.725$ with p-value $0.276$, suggesting that the sphericity assumption is not violated. Similarly for \textit{Coffee Table} which has Mauchly's W of $0.655$ with p-value $0.139$ and \textit{Dining Table} that has Mauchly's W of $0.570$ with p-value $0.110$, also indicating the assumption holds. Therefore, sphericity assumptions are met for all task families and the ANOVA results are valid.

\vspace{0.5em}
Finally, we performed the paired-sample t-test (Table \ref{tab:postHocComparisons}) to identify which specific pairs of methods are different from each other after establishing overall differences from the ANOVA. For \textit{Study Desk}, \model shows significant improvements over Direct LLM Prediction by $1.477$ units and even larger improvements over End-to-end Diffusion Model by $2.600$ units. The improvement over our ablation, i.e., LLM + Diffusion baseline, is smaller by still statistically significant with p-value $< 0.01$. Similar patterns were observed in \textit{Coffee Table} and \textit{Dining Table} comparisons where \model consistently outperformed all baseline methods compared, as reflected by significant mean differences and very low p-values. In summary, \model consistently shows significant improvements over other models across all task families. Adjusted p-values (Holm method) maintained statistical significance, further confirming robustness of these results.

%% file: appendix/12-user_generated_sketch.tex
\subsection{How Can New Users Generate the Sketch?}

We demonstrate how an end-user can create the sketches for new task families using a dressing table example:

\vspace{0.5em}
\textbf{Preparation}: Initially, users identify common objects in the task family and consider typical arrangements. This preliminary step aids in understanding the task's logic and serves as a sanity check for the sketch. Users write the function specification for the task family and the logical steps in natural language.

\vspace{0.5em}
\textit{Example: Common objects on a dressing table include makeup, skincare products, hair products, jewelry, and accessories. These might be clustered based on function and arranged orderly within each cluster.}

\begin{lstlisting}[language=Python]
def dressing_table_layout(instruction, object_list):
    """
    Setting up the dressing table based on user instructions and a list of objects.
        
    Workflow:
    1. Determine the task-relevant information to guide the functional arrangement of objects.
    2. Categorize objects into functional clusters based on their types like skincare, makeup, jewelry, etc.
    3. Develop and apply strategies for arranging items within each cluster.
    4. Establish rules for placing these clusters in relation to each other on the dressing table.
        
    Parameters:
    - instruction (str): Instruction for setting up, highlighting any special preferences.
    - object_list (list): Available objects for arrangement, denoted by their names.
        
    Returns:
    - List of active relationships delineating the arrangement of objects.
    """
\end{lstlisting}

\vspace{0.5em}
\textbf{Step 1: Task-Relevant Information}: Users determine which information significantly impacts functional object arrangement (FORM). This involves writing function specifications to query this information.

\vspace{0.5em}
\textit{Example: On a dressing table, relevant information might include the number of containers, which dictates how many functional clusters can be formed.}

\begin{lstlisting}[language=Python]
     # Step 1: Gather task-relevant information based on instructions
    containers_count = infer_containers_count(instruction, object_list)

\end{lstlisting}

\vspace{0.5em}
\textbf{Step 2: Functional Clustering of Objects}: Define key categories for clustering objects for FORM. Write function specifications to classify input objects into these categories.

\vspace{0.5em}
\textit{Example: Functional clusters might include skincare, makeup, jewelry, and accessories. Depending on the number of containers identified in Step 1, some categories might be combined.}

\begin{lstlisting}[language=Python]
    # Step 2: Create functional clusters based on object types
    skincare_items, makeup_items, jewelry_items, accessory_items = classify_into_clusters(object_list)
\end{lstlisting}

\vspace{0.5em}
\textbf{Step 3: Arranging Objects Within Clusters}: Write function specifications for arranging objects within each cluster.

\vspace{0.5em}
\textit{Example: for each functional cluster, such as skincare and makeups, we write a function name for the object arrangement within each cluster.}

\begin{lstlisting}[language=Python]
    # Step 3: Arrange objects within each cluster
    skincare_layout = arrange_skincare_items(skincare_items, instruction)
    makeup_layout = arrange_makeup_items(makeup_items, instruction)
    jewelry_layout = arrange_jewelry_items(jewelry_items, instruction)
    accessories_layout = arrange_accessory_items(accessory_items, instruction)
\end{lstlisting}

\vspace{0.5em}
\textbf{Step 4: Arranging Objects Among Clusters}: Establish function specifications for arranging objects among different clusters.

\vspace{0.5em}
\textit{Example: Typically, jewelry and other accessories are placed adjacent to each other; a function can be written to define the abstract relationships between these clusters.}

\begin{lstlisting}[language=Python]
    # Step 4: Define the arrangement among clusters
    active_relationships = arrange_clusters_on_table(skincare_layout, makeup_layout, jewelry_layout, accessories_layout, instruction)
\end{lstlisting}

\vspace{0.5em}
As the next step, we plan to leverage LLM capabilities to translate natural language descriptions directly into practical sketches, simplifying the process further.